\pdfoutput=1

\documentclass[11pt]{article}

\usepackage[preprint]{acl}

\usepackage{times}
\usepackage{latexsym}

\usepackage[T1]{fontenc}

\usepackage[utf8]{inputenc}

\usepackage{microtype}
\usepackage{adjustbox}
\usepackage{colortbl}
\usepackage{pgf}
\usepackage{xcolor}
\usepackage{makecell}

\usepackage{tabularx}

\usepackage{inconsolata}

\usepackage{graphicx}

\usepackage{tikz}
\usetikzlibrary{shapes.callouts, decorations.pathreplacing, positioning}
\usepackage{geometry}

\usepackage{algorithm}
\usepackage{algorithmicx}
\usepackage[linesnumbered,ruled,vlined,algo2e]{algorithm2e}
\usepackage{algpseudocode}
\usepackage{amsmath}
\usepackage{amsthm}
\usepackage{array}
\usepackage{booktabs}
\usepackage{calc}
\usepackage{colortbl}
\usepackage{enumitem}
\usepackage{footnote}
\usepackage{graphicx}
\usepackage{hyperref}
\usepackage{listings}
\usepackage{mathtools}
\usepackage{microtype}
\usepackage{multirow}
\usepackage{newfloat}
\usepackage{seqsplit}
\usepackage{subcaption}  
\usepackage{url}
\usepackage{xcolor}
\usepackage{xspace}
\usepackage{lineno}

\newcommand{\comment}[1]{\noindent\textit{//~#1}\par}
\newcommand{\dataset}{MHSD}

\newcommand{\method}{MedAgent\xspace}
\newcommand{\eval}{MultiSenseEval\xspace}

\definecolor{lowcolor}{RGB}{219, 129, 123}   
\definecolor{highcolor}{RGB}{67,160,71}   

\newcommand{\cellcolorgradient}[1]{%
  \pgfmathsetmacro{\raw}{#1}%
  \pgfmathparse{%
    \raw<=1  ? \raw*4 :      
    (\raw>4 ? \raw/3.5 :     
               \raw)         
  }%
  \let\val\pgfmathresult
  \pgfmathsetmacro{\val}{max(min(\val,4),1)}%
  \pgfmathtruncatemacro{\zone}{ifthenelse(\val<2.5,0,1)}%
  \ifcase\zone
    \pgfmathsetmacro{\intensity}{100 - (\val - 1)*50}%
    \cellcolor{lowcolor!\intensity!white}%
  \or
    \pgfmathsetmacro{\intensity}{20 + (\val - 2.5)*60}%
    \cellcolor{highcolor!\intensity!white}%
  \fi
  #1%
}

\title{\textit{Reasoning Is Not All You Need:} Examining LLMs for Multi-Turn Mental Health Conversations}

\author{Mohit Chandra$^{1}$\thanks{Equal contribution.}, Siddharth Sriraman$^{1}$\footnotemark[1], Harneet Singh Khanuja$^{1}$, \\
\textbf{Yiqiao Jin$^{1}$, Munmun De Choudhury$^{1}$} \\
  $^{1}$Georgia Institute of Technology \\
  \texttt{\{mchandra9,sidsr,hkhanuja3,yjin328\}@gatech.edu,munmun.choudhury@cc.gatech.edu} 
}

\begin{document}
\maketitle
\begin{abstract}
Limited access to mental healthcare, extended wait times, and increasing capabilities of Large Language Models (LLMs) has led individuals to turn to LLMs for fulfilling their mental health needs. However, examining the multi-turn mental health conversation capabilities of LLMs remains under-explored. Existing evaluation frameworks typically focus on diagnostic accuracy and win-rates and often overlook alignment with patient-specific goals, values, and personalities required for meaningful conversations. To address this, we introduce~\method, a novel framework for synthetically generating realistic, multi-turn mental health sensemaking conversations and use it to create the Mental Health Sensemaking Dialogue (MHSD) dataset, comprising over 2,200 patient–LLM conversations. Additionally, we present~\eval, a holistic framework to evaluate the multi-turn conversation abilities of LLMs in healthcare settings using human-centric criteria. Our findings reveal that frontier reasoning models yield below-par performance for patient-centric communication and struggle at advanced diagnostic capabilities with average score of $\sim$31\%. Additionally, we observed variation in model performance based on patient's persona and performance drop with increasing turns in the conversation. Our work provides a comprehensive synthetic data generation framework, a dataset and evaluation framework for assessing LLMs in multi-turn mental health conversations.
\end{abstract}

\section{Introduction}

With nearly 70\% of individuals around the globe having limited to no access to mental healthcare~\cite{kazdin2013novel} and wait times for new patients extending up to three months~\cite{apasurvey23}, traditional mental health care system is increasingly unable to meet the rising demand. At the same time, recent advancements in reasoning capabilities of LLMs has demonstrated significant performance improvement on challenging tasks~\cite{openaio1, guo2025deepseek, tu2025towards}. This has led individuals to turn to LLMs to fill the gap in access to mental healthcare by allowing them to understand their personal mental health situations~\cite{aydin2024large}. This process of sensemaking\footnote{Sensemaking refers to ``\textit{processes of interpretation and meaning production whereby individuals and groups interpret and reflect on phenomena}''~\cite{brown2008making}} encompasses a wide range of use-cases, from interpreting medical information to obtaining lifestyle recommendations and getting answers to health related questions. However, it remains unclear if LLMs can engage in meaningful multi-turn sensemaking conversations, especially pertaining to mental health conditions.

\begin{figure*}[!t]
    \centering
    \includegraphics[width=\linewidth]{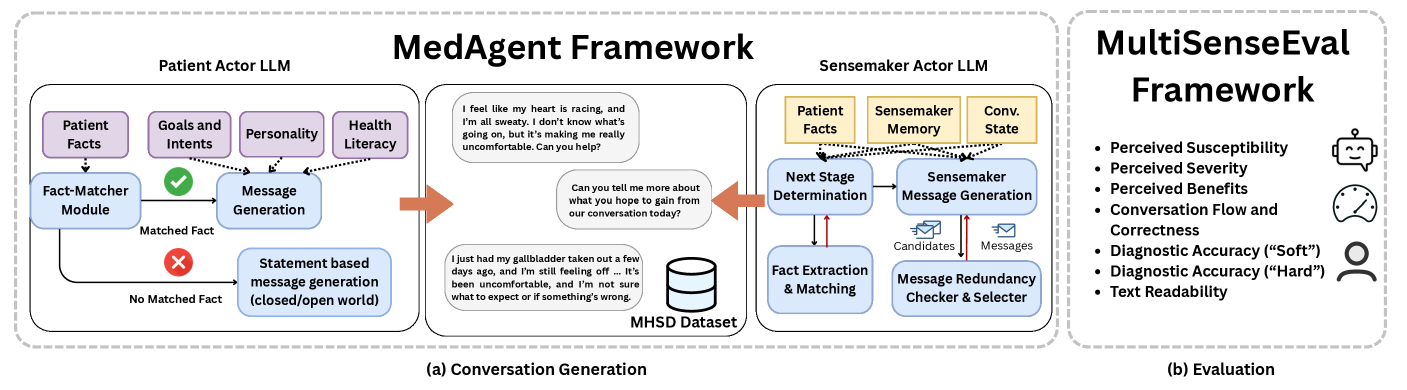}
    \caption{We present the~\method framework for generating realistic multi-turn mental health sensemaking conversations (\textbf{part (a)}). Using this framework we create the~\dataset~dataset with 2,284 conversations. Finally, we also present the~\eval framework (\textbf{part (b)}) to holistically evaluate LLM performance across patient-centric communication, conversational flow and correctness, diagnostic accuracy, and readability.}
    \label{fig:overview_figure}
    \vspace{-3mm}
\end{figure*}

Single-turn settings that are often used for assessing LLM performance in healthcare domain do not portray the real-world complexities that often require iterative information gathering, proactive follow-ups, and shared decision-making~\cite{dahm2022more, trevena2006systematic}. Recent works assessing multi-turn conversational capabilities of LLMs for general-purpose clinical conversations have reported significant decline in LLM performance for multi-turn dialogues~\cite{li2024mediq, liu2025exploring}.  However, such works have predominantly focused on the information gathering and diagnostic capabilities on general healthcare scenarios, often limiting evaluation to the point of diagnosis and reducing conversational capability analysis to classification or win-rate tasks. Filling this gap requires us to assess the capabilities of LLMs towards supporting multi-turn sensemaking conversations with the users through a more holistic approach taking account of patient-centric objective, values and needs. A major challenge in this direction is the limited to no access to real-world patient–clinician mental-health related conversation data the encompasses the full arc of interactions typically seen in clinical settings (such as the \emph{6-function model} of medical communication~\citep{de2009endpoints, king2013best}). Hence, our study is guided by the following research questions:
\vspace{0.5em}

\noindent\textbf{RQ1}: How can we effectively generate synthetic multi-turn mental health related sensemaking conversations between patients and LLMs?

\noindent\textbf{RQ2}: How well do reasoning LLMs conduct multi-turn mental health sensemaking conversations, and how does the performance change by patient persona and conversation length?
\vspace{0.5em}

To address these questions, we introduce~\method, a novel framework for synthetically generating multi-turn mental health conversations that are grounded in medical literature and portray real-world settings. \method is model-agnostic and can be directly used with any LLM. Using \method, we generate the~\textbf{M}ental \textbf{H}ealth \textbf{S}ense-making \textbf{D}ialogue (\dataset) dataset with 2,284 synthetic conversations between patients and LLMs. Finally, we introduce~\eval framework for holistically assessing the multi-turn sensemaking conversations focusing on six axes: (1) \emph{Perceived Susceptibility}, (2) \emph{Perceived Severity}, (3) \emph{Perceived Benefits}, (4) \emph{Diagnostic Accuracy}, (5)\emph{Conversation flow and Correctness}, and (6) \emph{Text readability}.

Our findings reveal limited capabilities of frontier reasoning models towards patient-centric communication and diagnostic accuracy. In our evaluation, OpenAI o1 and DeepSeek-R1 obtained an average score of 2.55 and 2.77 (on a scale of 4) on patient-centric communication metrics and an accuracy score of~$\sim$ 31\% for exact diagnosis matching, highlighting the below-par performance for both models across evaluation axes. Our study also revealed, disparity in model performance based on patient actor's persona, with models performing better for patients with `Agreeable' personality. Finally, we also observed that model performance for patient-centric communication and diagnostic metrics dropped with increasing number of turns in the conversation. Specifically, R1 exhibited a drop of  12.83\%, 6.71\%, and 29.6\% for \emph{Perceived Susceptibility}, \emph{Perceived Severity}, and \emph{Diagnostic Accuracy `Hard'} when the number of sensemaker messages increased from the 5
10-15. The proposed~\eval framework provides a more holistic approach towards assessing the quality of multi-turn mental health conversations. Additionally, recognizing community's need for multi-turn conversation data, we will publicly release the~\dataset~dataset.

\section{\method Framework}
\label{sec:conversation_generation_framework}

Figure~\ref{fig:overview_figure} (a) presents the overview of~\method framework.~\method framework generates mental health related sensemaking conversations using two actor LLMs: (1) Patient Actor LLM, and (2) Sensemaker Actor LLM. We use GPT-4o~\cite{gpt4o} for the patient actor LLM, and frontier reasoning models, OpenAI o1~\cite{openaio1} and DeepSeek-R1~\cite{guo2025deepseek} respectively for the sensemaker actor LLM. For the sensemaker actor llm, we specifically used reasoning models as its processes (such as moving stages in the conversation, validating hypothesis) require advanced reasoning capabilities. In the following subsections we describe the details for both actor components of the~\method framework.

\subsection{Patient Actor LLM}

Real-world sensemaking conversations are influenced by patient-specific factors such as medical background, personality traits, health literacy, individual goals and intentions~\citep{brown2008making,helms2010making}. To reflect this diversity, we constructed diverse patient personas for the simulated patient actor LLM by creating sets of patient's medical background, personality traits, health literacy, individual high-level intentions and concrete goals. Below, we describe each component of the patient actor LLM. Algorithm~\ref{alg:patient} presents the algorithm for generating the patient actor LLM message $p_t$ at timestamp \emph{t} based on the sensemaker message ($s_{t-1}$), patient persona $\Pi$, and set of patient's atomic facts $F$.

\vspace{0.1em}
\noindent\textbf{Patient Background}: We began with curating a seed set of clinical case studies to serve for generating simulated patient dialogues. We used MedQA~\cite{jin2021disease}, to obtain US medical board exam case-studies and filtered 181 case studies focusing on patients with mental health-related conditions using a GPT-4o-based classifier followed by human validation. Building on past work~\cite{min-etal-2023-factscore}, we further decomposed each case study into a list of atomic medical facts and also extracted the corresponding diagnostic labels. These atomic facts serve as the foundation for constructing the final patient persona.

\vspace{0.1em}
\noindent\textbf{Personality Traits}: To capture variability in personality, we grounded the patient personality traits based on the Big Five Personality Traits~\cite{goldberg1993structure,saucier1994mini,cobb2012stability}, a psychological framework describing personality in terms of five broad dimensions: \emph{Openness}, \emph{Conscientiousness}, \emph{Extraversion}, \emph{Agreeableness}, and \emph{Neuroticism}. To ensure behavioral consistency, we varied each trait one at a time on a binary scale, as past studies have shown varying correlation between the traits~\citep{klimstra2013correlated, yu2021relationship, kang2023personality}, and random sampling of traits may lead to internally contradictory personalities. Hence, for each case study, we created five personality variations in which one of the five personality traits in tuned to `High' to modulate their conversational behavior. For instance, patients with `High' \emph{Agreeableness} tend to seek social harmony and use affiliative language. We have provided examples of other traits in Appendix~\ref{sec:appendix_big_5_personality}.

\vspace{0.1em}
\noindent\textbf{Health Literacy}: Health literacy is described as ``\textit{personal knowledge and competencies which enable people to access, understand, appraise, and use information and services in ways which promote and maintain good health and well-being for themselves and those around them}''~\cite{nutbeam2021health}. Individuals with inadequate health literacy have been observed to have difficulty in participating in medical decision-making process, following medical recommendations and conversations~\cite{williams1995inadequate}. To account for such variations, we introduced two health literacy levels (\emph{Basic}, \emph{Advanced}) for the patient persona (detailed definitions and prompts in Appendix~\ref{sec:appendix_patient_health_literacy}).

\vspace{0.1em}
\noindent\textbf{Patient Intentions}: Past works have demonstrated that LLMs can support various patient needs and intentions related to patient education and sensemaking~\cite{10.1001/jamanetworkopen.2024.0357, bragazzi2024assessing, aydin2024large}. However, the patient backgrounds obtained from the MedQA dataset focused on diagnosis and lacked explicit representations of such intentions. To address this, we augmented each patient profile with realistic, intention-driven motivations for engaging with LLMs in sensemaking conversations. Building upon~\citet{aydin2024large} and~\citet{lipersonalized2024}, we used three different intentions (detailed definitions and examples are provided in Appendix~\ref{sec:appendix_patient_intentions}). To operationalize this augmentation, we employed GPT-4o to assign the most realistic and logical intention from one of the three intentions to the given patient background (prompt in Table~\ref{tab:appendix_patient_intent_augmentation_table}).

\vspace{0.1em}
\noindent\textbf{Patient Goals}: Patient goals refer to what they want to have achieved at the end of the conversation. However, similar to patient intentions, patient backgrounds lacked concrete goals for the conversations. To address this, we used three broad categories of goals as mentioned in~\citet{auriemma2024classification}, (1) \emph{Comfort-focused}, (2) \emph{Maintain or improve function}, and (3) \emph{Life extension} (detailed definitions and examples in Appendix~\ref{sec:appendix_patient_goals}). These goals focus on three different aspects patient-oriented care. For instance, while the \emph{Comfort-focused} goal aims to seek interventions to promote comfort and avoiding those that would increase discomfort, \emph{Life extension} deals with the patient's goal of extending longevity or survival even at the expense of maximizing function or comfort. Similar to the patient intention augmentation, we used GPT-4o to assign the most relevant and logical patient goal to the patient background (prompt in Table~\ref{tab:appendix_patient_goal_augmentation_table}).

\begin{algorithm}[!t]
\small
\caption{\textsc{PatientStep}$(s_{t-1},\;\Pi,\;\mathcal F)$}
\label{alg:patient}
\KwIn{Sensemaker utterance $s_{t-1}$, 
      patient profile $\Pi$,
      set of patient's atomic facts $\mathcal{F}$  }
\KwOut{$p_t$ \comment{patient reply at turn $t$}}
\SetKwFunction{MatchFacts}{MatchFacts}
\SetKwFunction{GenPatientMsg}{GenPatientMsg}
\SetKwFunction{RequestType}{RequestType}
\SetKwFunction{ReplyWithContext}{ReplyWithContext}
\SetKwFunction{ClosedWorldMessage}{ClosedWorldMessage}
\BlankLine
\comment{Match message mentioned by the sensemaker against patient facts.} 
$\mathcal{F}_{\mathrm{match}} \leftarrow$ \MatchFacts{$s_{t-1}$, $\mathcal{F}$} \\             
\comment{Check \#facts matched}
\If{$|\mathcal{F}_{\mathrm{match}}| > 0$}{
    \comment{Generate patient messages based on matched facts. Apply stylistic variations based on personality, health literacy, and goals/intents defined in $\Pi$.}
    $p_t \leftarrow$ \GenPatientMsg{$\mathcal{F}_{\mathrm{match}}$, $\Pi$}
}
\Else{
\comment{Determine whether the requested information is background information}
\uIf{\RequestType{$s_{t-1}$} $==$ background}{ 
    \comment{Apply closed world assumption. Indicating lack of information.}
    $p_t \leftarrow$ \ClosedWorldMessage{}
}\Else{
    $p_t \leftarrow$ \ReplyWithContext{$s_{t-1}$, $\Pi$} \comment{reply using general context and style}
}
}
\Return{$p_t$}
\end{algorithm}

\vspace{0.1em}
\noindent\textbf{Patient Actor LLM Message Generation Pipeline}: Algorithm~\ref{alg:patient} presents the algorithm for generating the patient actor LLM message $p_t$ at timestamp \emph{t} based on the sensemaker message ($s_{t-1}$), patient persona $\Pi$, and set of patient's atomic facts $F$. While patient persona $\Pi$ refers to the set of personality traits, health literacy level, patient intent and goals, patient's atomic facts $F$ are obtained from the patient background case study as described above. For a given timestamp \emph{t}, we first retrieve the relevant facts from the patient's atomic fact list $F$ that could be used to generate the reply using a fact matching and retrieval prompt (Appendix Table~\ref{tab:appendix_patient_answer_fact_selection_prompt_table}). If the number of facts matched ($|\mathcal F_{\mathrm{match}}|>0$) then we use the patient message generation module to generate the final message $p_t$ by combining the matched facts with stylistic variations based on personality, health literacy, and goals/intents defined in $\Pi$. In case, $|\mathcal F_{\mathrm{match}}|=0$, we generate messages based on the type of information being asked (either background or non-background). Background information here refers to the information which is related to the patient's medical facts obtained from the case study, on the other hand, non-background information refers to knowledge about future steps which are not defined in patient's background, intentions or goals. For background related information we assume a \emph{closed world} assumption and reply signifying that the patient is unsure about the required information. In case the information being requested is related to non-background aspects then patient module can generate an affirmative message based on the situation. The starting message $p_0$ is generated through a special prompt that uses the two most relevant case study facts, personality and the intent (prompt in Table~\ref{tab:appendix_patient_start_message_generation_table}).
\subsection{Sensemaker Actor LLM}
\label{subsec:sensemaker_actor_llm}

The goal of the sensemaker actor LLM is to produce informative, non-redundant, and stage-appropriate responses that emulate the structure of clinical conversations. Figure~\ref{fig:overview_figure} (b) shows the high-level sensemaker message generation pipeline, which includes four modules: (1) Next Stage Determination, (2) Fact Extraction \& Matching, (3) Sensemaker Message Generation, and (4) Message Redundancy Checker \& Selector. One of the novel features in the generated conversations is the existence of a structured organization of the sensemaking conversation into five stages inspired by \emph{6-function model} of medical communication~\citep{de2009endpoints, king2013best}. Hence, we first describe the stages of the sensemaking conversation followed by the overall working of the sensemaker actor LLM.

\vspace{0.1em}
\noindent\textbf{Stages of sensemaking conversation}: The sensemaker actor LLM organizes each conversation grounded within the \emph{6-function model} of medical communication~\citep{de2009endpoints, king2013best} (example conversations for each stage are in Appendix Table~\ref{tab:example_conversation}). The \emph{Fostering the Relationship} stage initiates the encounter by building trust, empathy, and psychological safety. The \emph{Gathering Information} stage focuses on eliciting both biomedical and psychosocial context through open-ended questions and active listening, aligning with how human physicians derive diagnostic hypotheses. In the \emph{Providing Information} stage, the sensemaker provides its diagnosis and explains it to the patient. The \emph{Decision Making} stage facilitates shared planning by integrating patient preferences and values. We combined the \emph{Enabling Disease and Treatment-related Behavior} stage with the \emph{Decision Making} stage due to the similar objectives within our context. Finally, the \emph{Responding to Emotions} stage enables agents to detect and attend to emotional distress. This phase-wise design not only mirrors real clinical workflows but also allows the sensemaker LLM to specialize in communication tasks that directly impact diagnostic accuracy, treatment adherence, and health outcomes. This design ensures that the sensemaker LLM is able to meaningfully engage in a multi-turn conversation. 

\begin{algorithm}[!h]
\small
\caption{Sensemaker Message Generation at Turn $t$}
\label{alg:sense_maker}
\KwIn{patient message $p_{t-1}$, known facts $F_{t-1}$, sensemaker message history $H$}
\KwOut{Next sense‑maker utterance $s_t$, updated facts $F_t$, stage index $g_t$}

\SetKwFunction{ExtractFacts}{ExtractFacts}
\SetKwFunction{MatchFacts}{MatchFacts}
\SetKwFunction{GetNextStage}{GetNextStage}
\SetKwFunction{Generate}{GenerateCandidates}
\SetKwFunction{Filter}{Filter}
\SetKwFunction{Sample}{Sample}

\textbf{Step 1: Stage Determination \& Fact Update} \\
\comment{Extract atomic facts from patient message} 
$\Delta F_t \leftarrow$ \ExtractFacts{$p_{t-1}$} \\
\comment{Merge with existing facts}
$\,\,F_t \leftarrow F_{t-1} \cup$ \MatchFacts{$\Delta F_t, F_{t-1}$} \\
\comment{Determine whether to move to next stages}
$\tilde{g}_t \leftarrow$ \GetNextStage{$g_{t-1}, p_{t-1}, F_t$} \\
\If{\#turns in current stage $g_{t-1} \ge 5$}{
    $g_t \leftarrow g_{t-1}+1$ \comment{Move to the next stage}
}
\Else{
    $g_t \leftarrow \tilde g_t$ \comment{Assign the computed next stage}
}

\textbf{Step 2: Candidate Generation} \\
\comment{Generate $|C|=3$ candidate messages}
$C \leftarrow$ \Generate{$g_t, F_t, H$} 

\textbf{Step 3: Redundancy Check} \\
\comment{Check each candidate message against past messages for redundancy. }
$C' \leftarrow$ \Filter{$C, H$}

\If{$|C'| > 0$}{
    $s_t \leftarrow$ \Sample{$C'$} \comment{sample one reply} \;
    $H \leftarrow H \cup \{s_t\}$
}
\Else{
    $g_t \leftarrow g_t + 1$ \comment{avoid stagnation}
}
\Return{$(s_t, F_t, g_t)$}
\end{algorithm}

\vspace{0.1em}
\noindent\textbf{Sensemaker Actor LLM Message Generation Pipeline}: Algorithm~\ref{alg:sense_maker} presents the algorithm for generating message $s_t$ at timestamp \emph{t} given the patient message $p_{t-1}$, known facts about the patients until timestamp~\emph{t} ($F_{t-1}$), and sensemaker message history $H$. After the sensemaker receives the patient's message $p_{t-1}$, it first invokes the \emph{Next Stage Determination} module. This module takes as input the current message $p_{t-1}$, known patient facts from the previous stages $F_{t-1}$, and the sensemaker's previous messages $H$ and passes $p_{t-1}$ to \emph{Fact Extraction and Matching}, which extracts new atomic facts $\Delta t$ and update the fact set $F_t$. Based on the updated facts $F_t$ and sensemaker's previous message history $H$, the \emph{Next Stage Determination} module decides whether to remain in the current stage of the conversation or advance to the next stage of the conversation. 
The information about the current stage $g_t$, patient facts $F_t$, and sensemaker message history $H$ are passed to the \emph{Sensemaker Message Generation} module. This module then produces $3$ candidate messages. The candidate messages are then sent to the \emph{Message Selection and Redundancy Checking} module, which filters out replies that are semantically similar to prior sensemaker messages ($H$). If non-redundant messages exist, one of such message is randomly selected as the output $s_t$. Otherwise, the conversation moves to the next stage. (Refer to Appendix \ref{sec:appendix_sensemaker_next_stage_determination_prompt}, \ref{sec:appendix_sensemaker_fact_extraction_and_matching_prompt}, \ref{sec:appendix_sensemaker_message_generation_prompts} and \ref{sec:appendix_sensemaker_redundancy_checking_prompt} for prompt details).

\section{\dataset~Dataset}
\label{sec:dataset_details}

\begin{figure}[!h]
\footnotesize 
\centering
\scalebox{0.5}{
\begin{tikzpicture}[
    node distance=0.5cm, every node/.style={inner sep=5pt, font=\itshape}, text width=7cm,
    brace/.style={decorate, decoration={brace, amplitude=10pt}},
    label/.style={font=\normalfont, align=left}
]


\node[font=\bfseries, align=center] at (-1.5,4) {Patient Actor};
\node[font=\bfseries, align=center] at (4.9,4) {Sensemaker};

\node[inner sep=0pt] at (-1.68,2.95) {
    \includegraphics[width=1cm]{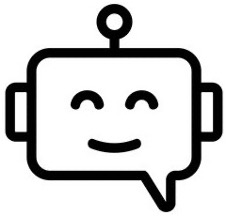}
};

\node at (4.2, 1.7) {
  \begin{minipage}{2cm}
    \centering
    \includegraphics[width=1cm]{figures/LLM_Icon.jpeg}
  \end{minipage}
};

\node[draw, fill=gray!10!blue!5, align=left, 
  callout pointer width=6pt,
  callout relative pointer={(0.1,-1)}, 
  rounded corners=5pt] 
(patient1) at (-1.6,1.5) {``My mom says I’ve been causing trouble in class. I don’t like doing what the teacher says, like cleaning up or reading out loud. I also don’t want to do gym class, but I still play the same games at recess.''};

\node[above=0.1cm of patient1,
      draw=none, 
      fill=gray!15,
      font=\color{black},
      align=left, 
      text centered, 
      text width=3.5cm,
      inner sep=2pt] 
(stage1) {O: $\downarrow$ C: $\uparrow$ E: $\downarrow$ A: $\downarrow$ N: $\downarrow$ Medical Literacy: Basic};

\node[draw, rounded corners, fill=gray!10!blue!5, 
      callout relative pointer={(-0.5cm,-0.5cm)}, align=left, 
      right=of patient1, yshift=-0.2cm] 
(Sensemaker1) {"You mentioned not wanting to do gym class but still playing games at recess—what feels different about recess compared to gym time?''};

\node[above=0.0cm of Sensemaker1,
      draw=none, 
      fill=gray!10!blue!5,
      font=\color{black},
      align=left, 
      text centered, 
      text width=3.7cm,
      inner sep=2pt] 
(stage1) {Fostering The Relationship};

\node[draw, rounded corners, fill=gray!10!blue!5, 
      callout relative pointer={(0.5cm,-0.5cm)}, align=left, 
      below=of patient1] 
(patient2) {"I don’t know, I guess recess just feels more relaxed. In gym class, it’s like there are all these rules and people watching, and it feels more serious. At recess, I can just play without worrying about all that stuff. It’s not as... I don’t know, strict?''};

\node[draw, rounded corners, fill=red!10!white, 
      callout relative pointer={(-0.5cm,-0.5cm)}, align=left, 
      right=of patient2, yshift=-0.2cm] 
(Sensemaker2) {"When gym class gets stressful, do you notice any physical changes—like your heart racing or your stomach feeling tight?''};

\node[above=0.0cm of Sensemaker2, 
      draw=none, 
      fill=red!10!white, 
      font=\color{black}, 
      text centered, 
      text width=3cm,
      inner sep=2pt] 
(stage2) {Gathering Information};

\node[draw, rounded corners, fill=red!10!white, 
      callout relative pointer={(0.5cm,-0.5cm)}, align=left, 
      below=of patient2] 
(patient3) {"No, I haven’t noticed any physical changes like that during gym class.'' \textsc{\small (closed world assumption)}};

\node[draw, rounded corners, fill=red!10!white, 
      callout relative pointer={(-0.5cm,-0.5cm)}, align=left, 
      right=of patient3, yshift=-0.2cm] 
(Sensemaker3) {"When you feel watched during gym class, what emotions come up for you—is it more nervousness, frustration, or something else?''};

\node[draw, rounded corners, fill=red!10!white, 
      callout relative pointer={(0.5cm,-0.5cm)}, align=left, 
      below=of patient3] 
(patient4) {"Yes, I’d say it’s more frustration than anything else. It just makes me feel uncomfortable, like I’m being judged or something.''};

\node[draw, rounded corners, fill=violet!15!white, 
      callout relative pointer={(-0.5cm,-0.5cm)}, align=left, 
      right=of patient4, yshift=-1cm] 
(Sensemaker4) {"Based on what you’ve shared, your frustration and discomfort in structured settings like gym class may point to \textbf{Oppositional Defiant Disorder}, where strict rules or perceived judgment trigger resistance.''};

\node[above=0cm of Sensemaker4, 
      draw=none, 
      fill=violet!15!white,
      font=\color{black},
      text centered, 
      text width=3cm,
      inner sep=2pt] 
(stage3) {Providing Information};

\node[draw, rounded corners, fill=violet!15!white, 
      callout relative pointer={(0.5cm,-0.5cm)}, align=left, 
      below=of patient4] 
(patient5) {"Yeah, I guess that makes sense. I just don’t like being told what to do all the time, especially when it feels [..] I don’t want to do. It’s annoying, and I don’t see why I have to follow all these rules.''};

\node[draw, rounded corners, fill=green!10!white, 
      callout relative pointer={(-0.5cm,-0.5cm)}, align=left, 
      right=of patient5, yshift=-1.3cm] 
(Sensemaker5) {"What if we practiced ways to communicate your needs in these situations—like requesting to do warm-ups your way first before joining group activities?''};

\node[above=0cm of Sensemaker5, 
      draw=none, 
      fill=green!10!white,
      font=\color{black},
      text centered, 
      text width=2.3cm,
      inner sep=2pt] 
(stage4) {Decision Making};

\node[draw, rounded corners, fill=green!10!white, 
      callout relative pointer={(0.5cm,-0.5cm)}, align=left, 
      below=of patient5] 
(patient6) {"I guess that could work. If I could start off doing things my way, maybe it wouldn’t feel so... I don’t know, uncomfortable or forced. I don’t really like jumping into group stuff [...] make it easier to join in after.'' \textsc{\small (low extraversion)}};

\node[draw, rounded corners, fill=orange!15!white,
      callout relative pointer={(-0.5cm,-0.5cm)}, align=left, 
      right=of patient6, yshift=-1cm] 
(Sensemaker6) {"It’s okay to need space to transition—if you try this approach, how confident do you feel about communicating it to the teacher when gym class starts?''};

\node[above=0cm of Sensemaker6, 
      draw=none, 
      fill=orange!15!white,
      font=\color{black},
      text centered, 
      text width=3.3cm,
      inner sep=2pt] 
(stage5) {Responding to Emotions};

\node[draw, rounded corners, fill=orange!15!white, 
      callout relative pointer={(0.5cm,-0.5cm)}, align=left, 
      below=of patient6] 
(patient7) {"I don’t feel very confident about talking to the teacher, honestly. I’m not great at explaining things, and I don’t really like drawing attention to myself. I’d rather just keep quiet and hope it works out on its own.''};

\node[right=of patient7, 
      draw=none,
      align=left,
      font=\color{black},
      text centered, 
      text width=3cm,
      inner sep=2pt, 
      yshift=0.6cm, xshift=-1cm] 
(stage3) {};

\node[draw, rounded corners, fill=gray!20,
      callout relative pointer={(-0.5cm,-0.5cm)}, align=left, 
      right=of patient7, yshift=-0.6cm] 
(Sensemaker8) {"You’ve made a great first step toward finding what works for you. Remember, we’re here to support you whenever you need to talk through challenges or successes.''};

\end{tikzpicture}
}
\caption{Sample conversation between the sensemaker and a patient with high conscientiousness (traits listed by first letter) and basic medical literacy. Stages are distinguished by color, with some intermediate dialogues skipped for conciseness. Diagnosis is in bold.}
\label{fig:sample_conversation}
\vspace{-2em}
\end{figure}

After initial post-processing to remove erroneous generations and incomplete conversations, we obtained 1,142 conversations each for DeepSeek-R1 and OpenAI o1 as the sensemaker in the~\dataset~ dataset (total conversations: 2,284). We used GPT-4o for all modules of the patient actor, and for all sensemaker modules excluding next stage determination and answer generation (which used the reasoning models). Table \ref{tab:conversation-metrics} presents a stage-wise breakdown of conversations. o1 conversations had 13.65 messages on average, while R1 conversations were longer, with 19.09 messages per conversation on average. o1 model also spent longer on gathering information from the patient before decision making (7.64 messages) compared to R1 (3.88 messages). An end-to-end sample conversation is shown in Figure \ref{fig:sample_conversation}.

\begin{table}[!h]
\centering
\small
\begin{tabular}{p{0.43\linewidth} p{0.2\linewidth} p{0.23\linewidth}}
\textbf{\small{Stage}} &  \multicolumn{2}{c}{\small{\textbf{Avg. Number of Messages}}} \\
 & {\small{OpenAI o1}} & {\small{DeepSeek-R1}} \\
\hline 
\small{Fostering the Relationship} & 1.11 & 1.22 \\
\small{Gathering Information} & 7.64 & 3.88 \\
\small{Providing Information} & 2.00 & 2.00 \\
\small{Decision Making} & 3.52 & 2.70 \\
\small{Responding to Emotions} & 3.83 & 2.86 \\
\small{Exit} & 1.00 & 1.00 \\
\hline
\small{Overall} & 19.09 & 13.65 \\
\end{tabular}
\caption{Stage-wise and overall average number of messages of the \dataset \hspace{.1cm}dataset with OpenAI o1 and DeepSeek-R1 as sensemaker, including both patient and sensemaker responses.
}
\label{tab:conversation-metrics}
\vspace{-2.5em}
\end{table}


\section{Sensemaking Evaluation}
\label{sec:evaluation_metrics}

In this section, we outline our approach for evaluating the sensemaking capabilities of LLMs. We evaluate the multi-turn conversations generated by two reasoning models: (1) OpenAI o1~\cite{openaio1}, and (2) DeepSeek-R1~\cite{guo2025deepseek} using a hybrid approach combining automated metrics and human evaluation. While automated evaluation allows us to evaluate the conversations on scale, human evaluation provides an validation of automated metrics.

\noindent\textbf{Automated Evaluation}: For automated evaluation we used a combination of multiple metrics assessing various aspects related to the sense-making conversation such as patient-centric communication, diagnostic accuracy, conversation flow and correctness, readability. For \emph{\textbf{patient-centric communication}}, we adopted metrics from the Health Belief Model (HBM)~\citep{champion2008health,janz1984health} -- a psychological framework that seeks to explain and predict individual health behaviors by examining individuals' attitudes and beliefs. Specifically, we use three dimensions from HBM namely perceived susceptibility, perceived severity, perceived benefits. To assess \emph{\textbf{diagnostic accuracy}}, we used ground truth diagnostic labels from the seed case studies obtained from~\citet{jin2021disease}. In contrast to the past works~\cite{li2024mediq, liu2025exploring}, we assessed diagnostic accuracy without providing multi-choice options to better reflect real-world conditions. Since, sensemaker's diagnosis is provided in a free-text generation, we used two different accuracy assessment settings: (1) Hard Diagnostic Accuracy, requiring exactly matching the sensemaker's diagnosis with the ground truth, and (2) Soft Diagnostic Accuracy which allowed for a broader category level matches for diagnosis.~\emph{\textbf{Conversation flow and Correctness}} measures the ability of sensemaker LLM to perform the end-to-end conversation maintaining the logical consistency, covering each phase of the conversation appropriately. Given, that the sensemaker LLM generates open-ended messages, we used LLM-as-a-judge approach to measure the aforementioned metrics. Finally, we measure the overall \emph{\textbf{readability}} of the responses provided by the sensemaker LLM using the SMOG index, a popular readability index to assess health literacy material.~\cite{mc1969smog}. Detailed information related to the definition, sub-criteria, evaluation rubric and prompts can be found in Appendix~\ref{sec:appendix_automated_conversation_evaluation}.

\noindent\textbf{Human Evaluation}: In addition to the automated evaluation, we also conducted human-evaluation as a validation check for automated metric scores on a subset of 100 conversations. Specifically, we used the same rubric as the automated evaluation for patient-centric communication, diagnostic accuracy, and conversation flow and correctness. We have provided details regarding the human evaluation in Appendix~\ref{appendix_sec:human_annotation_details}.
\section{Results}
\label{sec:results}

\subsection{\eval Framework Evaluation}

\begin{figure}[!h]
    \centering
    \includegraphics[width=0.8\columnwidth]{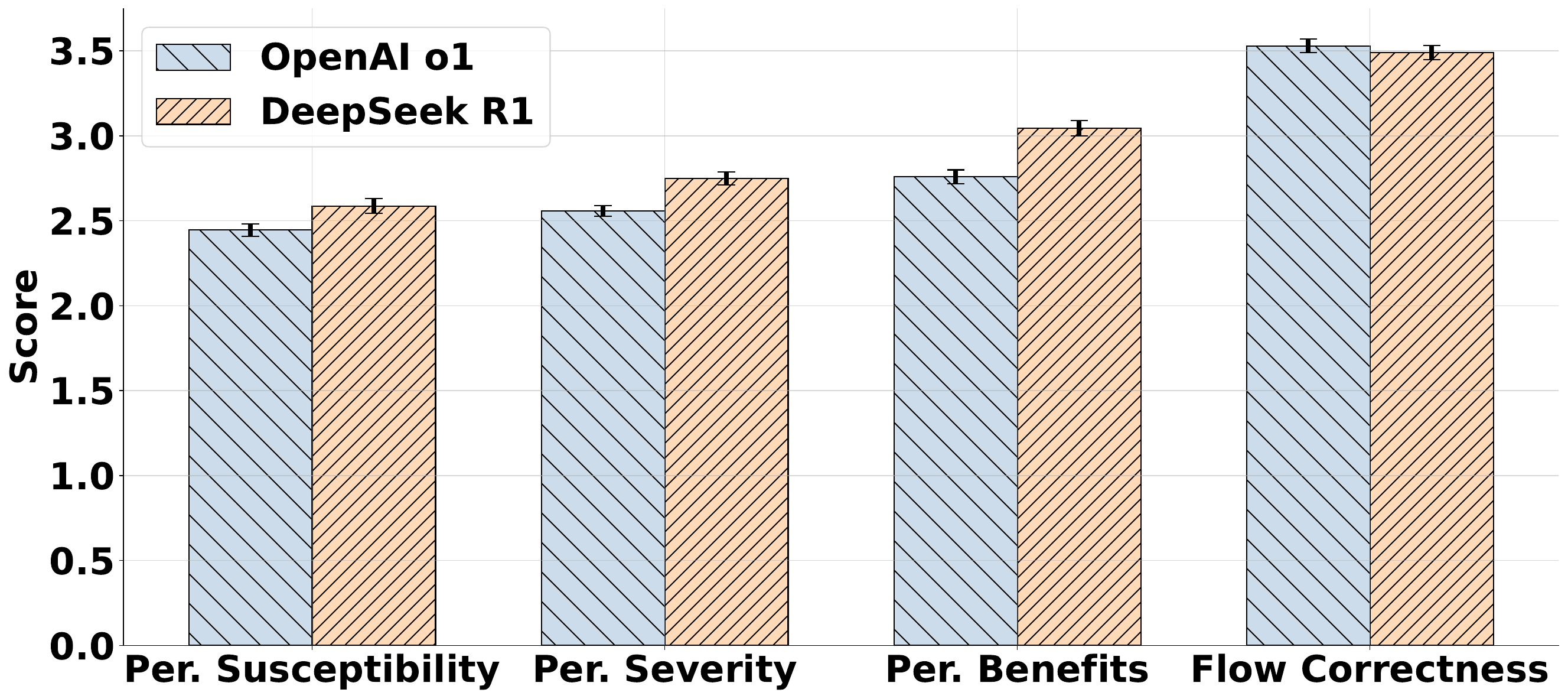}
    \caption{Performance comparison of OpenAI o1 and DeepSeek-R1 across~\textbf{Perceived Susceptibility},~\textbf{Perceived Severity},~\textbf{Perceived Benefits}, and~\textbf{Conversation Flow and Correctness}. Bars indicate mean scores with 95\% confidence interval. All scores are on a 4-point Likert scale ((1): \emph{Very Poor} to (4): \emph{Very Good}). Both models obtain scores below ``Good Performance'' rating for the three patient-centric communication metrics, but exceed the ``Good Performance'' threshold for Conversation Flow and Correctness.}
    \label{fig:metric_comparison_bar_graph_patient_centered}
    \vspace{-0.5em}
\end{figure}

\noindent\textbf{Patient-Centric Communication}: Assessing the patient-centric communication capabilities of OpenAI o1 and DeepSeek R1, we observed that R1 outperformed o1 across all three dimensions (Figure~\ref{fig:metric_comparison_bar_graph_patient_centered}). However, the mean average score for both models remained below than 3 on a 4-point scale (1: Very Poor to 4: Very Good) across all metrics, indicating the below-par performance towards appropriately communicating with the patient. For~\emph{Perceived Susceptibility}, R1 obtained a significantly higher average score of 2.59 (SD = 0.75) compared to score of 2.45 (SD = 0.63) obtained by o1 (Mann–Whitney U test; p$<0.001$; U=$597057$). Similarly, R1 obtained significantly higher scores for~\emph{Perceived Severity} (2.75; SD=0.67) compared to o1 (2.56; SD=0.54) (Mann–Whitney U test; p$<0.001$; U=$559539$). Lastly, R1 also obtained significantly higher score (3.04; SD=0.77) compared to o1 (2.76; SD=0.71) for the~\emph{Perceived Benefits} dimension (Mann–Whitney U test; p$<0.001$; U=$520950$). To qualitatively assess the shortcomings, we analyzed the lowest scoring conversations (score of 1 or 2) across all three metrics, revealing that reasoning models still lack personalization during mental health multi-turn conversations (detailed analysis in Appendix \ref{sec:appendix_qualitative_analysis}).

\noindent\textbf{Conversation Flow and Correctness}: Both OpenAI o1 and DeepSeek R1's scores exceeded the ``Good Performance'' threshold ($\geq3$ on a 4-point Likert scale), indicating that both models covered all five stages in the correct order with reasonable depth (as described in~\ref{subsec:sensemaker_actor_llm}). Interestingly, here we observed o1 outperforming R1 with models obtaining scores of 3.53 (SD=0.67) and 3.49 (SD=0.72) respectively (Figure~\ref{fig:metric_comparison_bar_graph_patient_centered}). Similarly for~\textbf{Text Readability}, we observed that
o1 produced significantly easier to read messages with a SMOG score of 11.50 (SD=1.26) compared to a score of 13.33 (SD=1.47) for DeepSeek R1 (t-test; p$<0.001$; t=$-31.95$). However, health organizations recommend SMOG scores between 6-8 ~\cite{badarudeen2010assessing}, which is significantly lower than the SMOG scores produced by both models.

\begin{figure}[!h]
    \centering
    \includegraphics[width=0.75\columnwidth]{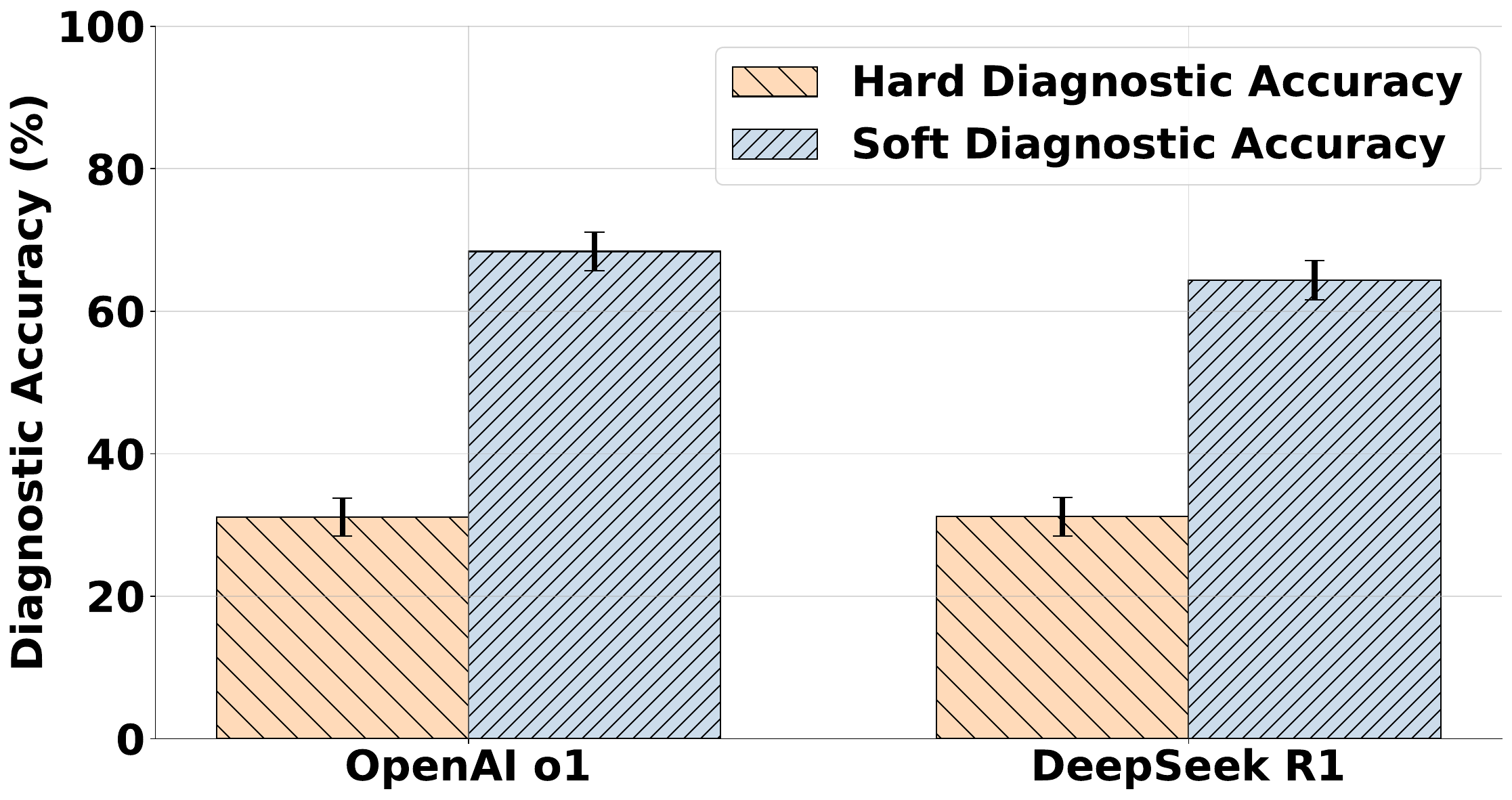}
    \caption{Performance comparison between OpenAI o1 and DeepSeek R1 across~\textbf{Hard Diagnostic Accuracy}, and ~\textbf{Soft Diagnostic Accuracy}. Bars indicate the mean scores with 95\% confidence interval. As observed the performance for both models drops by more than 50\% when the diagnosis is matched exactly with the ground truth (``Hard Accuracy'') in comparison to when it is matched on broader/general criteria (``Soft Accuracy'').}
    \label{fig:metric_comparison_bar_graph_diagnostic accuracy}
    \vspace{-1em}
\end{figure}

\begin{table*}[!h]
    \centering
  \begin{adjustbox}{width=\linewidth}
    \begin{tabular}{p{0.17\textwidth}|c|*{7}{|p{0.15\textwidth}}}
      \toprule
      \textbf{Personality Trait} & \textbf{Model}
  & \makecell{\textbf{Per.}\\\textbf{Susceptibility}}
  & \makecell{\textbf{Per.}\\\textbf{Severity}}
  & \makecell{\textbf{Per.}\\\textbf{Benefits}}
  & \makecell{\textbf{Flow}\\\textbf{Correctness}}
  & \makecell{\textbf{Diag. Acc.}\\\textbf{(Hard) (\%)}}
  & \makecell{\textbf{Diag. Acc.}\\\textbf{(Soft) (\%)}}
  & \textbf{SMOG}
      \\
      \midrule
      \multicolumn{9}{c}{\textbf{Basic Health Literacy Patient}} \\
      \midrule
      \multirow{2}{*}{Openness}
        & o1 & 2.46 & 2.62 & 2.91 & 3.53 & 31.68 & \underline{70.30} & 11.22 \\
        & R1 & 2.58 & 2.71 & \textbf{3.25} & 3.56 & \underline{34.65} & 58.42 & 13.27 \\
      \hline
      \multirow{2}{*}{Conscientiousness}
        & o1 & 2.45 & 2.53 & 2.77 & \underline{3.57} & 30.08 & 66.92 & 11.25 \\
        & R1 & 2.57 & 2.72 & \underline{3.08} & 3.47 & 29.32 & 64.66 & 13.23 \\
      \hline
      \multirow{2}{*}{Extraversion}
        & o1 & 2.44 & 2.50 & 2.69 & 3.47 & 25.49 & 66.67 & 11.23 \\
        & R1 & 2.44 & \underline{2.74} & 2.93 & 3.38 & 28.43 & 62.75 & 13.01 \\
      \hline
      \multirow{2}{*}{Agreeableness}
        & o1 & 2.42 & 2.44 & 2.85 & \textbf{3.68} & \textbf{36.43} & \textbf{70.54} & \underline{11.13} \\
        & R1 & \underline{2.60} & \underline{2.74} & 3.04 & 3.50 & 26.36 & 59.69 & 13.06 \\
      \hline
      \multirow{2}{*}{Neuroticism}
        & o1 & 2.19 & 2.43 & 2.41 & 3.24 & 31.33 & 67.47 & \textbf{10.92} \\
        & R1 & \textbf{2.66} & \textbf{2.81} & 2.88 & 3.28 & 28.92 & 59.04 & 12.90 \\
      \midrule
      \multicolumn{9}{c}{\textbf{Advanced Health Literacy Patient}} \\
      \midrule
      \multirow{2}{*}{Openness}
        & o1 & 2.56 & 2.63 & 2.82 & 3.59 & 29.37 & \textbf{72.22} & 11.88 \\
        & R1 & \underline{2.63} & \underline{2.79} & \underline{3.11} & \textbf{3.64} & 28.57 & 69.05 & 13.57 \\
      \hline
      \multirow{2}{*}{Conscientiousness}
        & o1 & 2.43 & 2.56 & 2.85 & 3.57 & 25.93 & 67.41 & 12.09 \\
        & R1 & 2.60 & 2.76 & 3.10 & 3.61 & \underline{38.52} & \underline{71.85} & 13.77 \\
      \hline
      \multirow{2}{*}{Extraversion}
        & o1 & 2.52 & 2.70 & 2.79 & 3.54 & 29.91 & 69.23 & \underline{11.65} \\
        & R1 & 2.59 & \underline{2.79} & 2.94 & 3.39 & 32.48 & 68.38 & 13.39 \\
      \hline
      \multirow{2}{*}{Agreeableness}
        & o1 & 2.50 & 2.57 & 2.84 & \underline{3.62} & 32.81 & 68.75 & 11.79 \\
        & R1 & \textbf{2.67} & \textbf{2.81} & \textbf{3.12} & 3.60 & 30.47 & 66.41 & 13.56 \\
      \hline
      \multirow{2}{*}{Neuroticism}
        & o1 & 2.39 & 2.57 & 2.47 & 3.30 & \textbf{39.77} & 62.50 & \textbf{11.53} \\
        & R1 & 2.48 & 2.60 & 2.89 & 3.32 & 34.09 & 57.95 & 13.31 \\
      \bottomrule
    \end{tabular}
  \end{adjustbox}
  \caption{Performance of o1 vs.\ R1 across seven metrics, by persona and health‐literacy level for three personality traits. We report the scores with the best and second-best performing model metrics in each scenario highlighted in \textbf{bold} and \underline{underline}, respectively. Dia. Acc. represents the Diagnostic Accuracy metric. }
  \label{tab:personality_based_analysis}
\end{table*}

\noindent\textbf{Diagnostic Accuracy}: For the ``Soft'' setting evaluation, o1 significantly outperformed R1 (o1: acc.score= 68.39\%; SD=46.52\%, R1: acc.score= 64.36\%; SD=47.91\%)(Mann–Whitney U test; p$<0.05$; U=$678348$) (Figure~\ref{fig:metric_comparison_bar_graph_diagnostic accuracy}). However, under the ``Hard'' setting, both models' performance dropped by over 50\%, with o1 obtaining an accuracy score of 31.09\% (SD=46.31\%) and R1 obtaining a score of 31.18\% (SD=46.34\%). These findings highlight the limited diagnostic capabilities of frontier reasoning models within multi-turn conversations requiring probing, consistent with prior findings~\cite{li2024mediq}.

\begin{figure*}[!h]
    \centering
    \includegraphics[width=\linewidth]{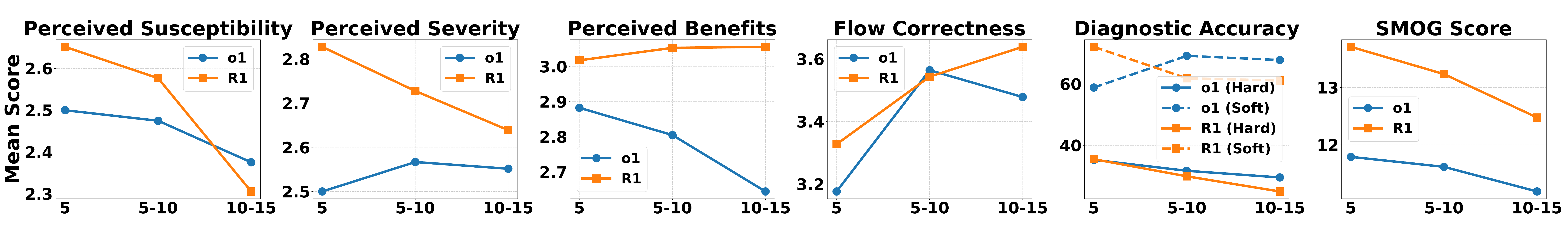}
    \caption{Performance trend for o1 and R1 across the~\eval framework metrics, the x-axis indicates sensemaker message count bins. While on average performance on patient-centric metrics and diagnostic accuracy declined with longer conversations, flow correctness and readability improved.}
    \label{fig:sensemaker_turn_count_performance_change}
    \vspace{-1em}
\end{figure*}

\subsection{Performance Disparity based on Patient Persona}

To understand how differences in patient persona may impact model's performance, we stratified the OpenAI o1 and DeepSeek R1's evaluation results across all performance metrics (ref. Section~\ref{sec:evaluation_metrics}) based on patient's dominant personality traits (Openness, Conscientiousness, Extraversion, Agreeableness, Neuroticism) and health literacy level (Basic/Advanced). Table~\ref{tab:personality_based_analysis} presents results across these strata for three personality types.

\noindent\textbf{Impact of Personality Traits}: Among patient with advanced health literacy, models performed best along the~\emph{Perceived Susceptibility}, \emph{Perceived Severity}, and \emph{Perceived Benefits} metrics for patient with dominant personality trait as `Agreeableness'. For instance, R1 achieved scores of 2.67 (susceptibility), 2.81 (severity), and 3.12 (benefits) for patients belonging to this cohort. In contrast, R1's performance dropped by 7.1\%, 7.5\%, and 7.4\% for the aforementioned metrics for patients with same literacy skills but with `Neuroticism' personality trait. We observed, a similar trend among patients with basic health literacy, where models performed better on \emph{Flow Correctness} and \emph{Diagnostic Accuracy} metrics for patients with `Agreeableness' personality. These findings reveal the performance gap towards patient-centric communication, where models performed better for patients with higher degree of prosocial and cooperative behavior (``Agreeableness'') compared to patients with other personality profiles. Interestingly, o1 generated easier to read messages for patients with `Neuroticism' dominant personality compared to other personality type. For such patients, the SMOG score for o1 responses were 2.7\% and 2.95\% lower compared to that for patients with `Openness' personality with basic and advanced literacy skills.

\noindent\textbf{Impact of Health Literacy}: Comparing model performance across patients with basic and advanced health literacy, we observed gain in o1 (+2.77\%) and R1's (+0.63\%) performance for patient-centric communication metrics. Interestingly we observed contrasting trends in performance, for o1 the accuracy decreased by 0.12\% for ``soft'' and 0.73\% ``hard'' diagnostic accuracy settings, whereas we observed an increase of 10.16\% and 11.74\% in ``soft'' and ``hard'' diagnostic accuracy for R1. Additionally, both models produced harder to read text when interacting with patient actors with higher literacy, with SMOG scores increasing by 5.84\% and 3.29\% for o1 and R1 respectively. These findings highlight a trade off in performance of models, while models may provide better patient-centric messages to patients with higher health literacy, those messages could be harder to read.

\subsection{Impact of Conversation Length on Performance}

We also examined the change in model performance with increased conversation length. Figure~\ref{fig:sensemaker_turn_count_performance_change} presents the line plots for o1 and R1 performance for each evaluation metric across number of sensemaker messages in the conversation. Interestingly, for R1, we observed that the performance decreased by 12.83\%, 6.71\%, and 29.6\%  for \emph{Perceived Susceptibility}, \emph{Perceived Severity}, and \emph{Diagnostic Accuracy (``Hard'')} respectively when the number of sensemaker messages increased from the 5 to 10-15. We observed a similar trend for o1 with a performance drop of 4.8\% and 7.99\% for \emph{Perceived Susceptibility} and \emph{Perceived Benefits} respectively, highlighting the reduced performance among both models for patient-centric communication with increased number of messages. In contrast, both models showed improvement in performance for the \emph{Conversation Flow Correctness} and \emph{Text Readability} performance with increased number of turns. These findings further highlight that even frontier reasoning models struggle to sustain effective performance over longer conversations, a finding highlighted in contemporary work~\cite{laban2025llms}.

\section{Related Work}

\noindent\textbf{LLM Evaluation for Single Healthcare Conversations}: Recent advancements in large language models (LLMs) have enabled diverse applications among 
clinicians~\cite{tu2025towards, singhal2023towards} and patients~\cite{yang-etal-2023-towards}. 
Models such as Med-PaLM~\citep{tu2024towards} and OpenAI's GPT series~\citep{chatgpt} have demonstrated strong performance on various medical benchmarks~\cite{nori2023capabilities, singhal2023large, singhal2023towards}, , highlighting their promise in clinical settings. However, critical challenges remain, including hallucinations~\cite{agarwal2024medhalu, asgari2025framework,deng2024deconstructing} cross-lingual disparities~\cite{jin2024better}, sociocultural biases (e.g. gender and geographic)~\cite{restrepo2024analyzing,liu2025culturevlm}, and misalignment with experts~\cite{chandra-etal-2025-lived}, and limitations in clinical competency tests
~\citep{thirunavukarasu2023trialling, jin2023psyeval}.

\vspace{0.5em}
\noindent\textbf{Multi-turn Conversation Evaluation for LLMs}: Beyond single-turn settings, recent works have also explored LLMs'multi-turn clinical conversations abilities. For example, MediQ~\citep{li2024mediq} evaluates question-asking ability of LLMs within multi-turn conversations. Along similar lines, researchers proposed a proactive dialogue generation framework based on dialogue ranking~\cite{li2025two}. The Ask Patients with Patience (APP) framework~\citep{zhu2025ask} allow LLMs to generate multi-turn conversations based on medical guidelines and entropy minimization. In contrast, ~\citep{liu2025exploring} proposed a patient simulator for multi-turn diagnostic conversations. However, past works overlook the diagnostic capabilities of LLMs in multi-turn conversations.

\vspace{-0.5em}
\section{Conclusion}

In this work, we proposed~\method, a framework for creating realistic mental health sensemaking conversations. Using~\method, we created~\dataset~dataset consisting of 2,284 synthetic sensemaker-patient conversations. Finally, we presented the~\eval framework for holistically evaluating multi-turn mental health conversation using metrics grounded in clinical research. Our findings reveal that frontier reasoning models yield below-par performance for patient-centric communication. Additionally, our study reveal the performance disparity among models based on patient's persona and reduction in model performance with increase number of turns in the conversation. Our work provides a comprehensive method for creating synthetic clinical conversations, a dataset and evaluation framework for assessing LLMs in multi-turn mental health conversations.

\section{Broader Implications and Potential Risks}

LLMs have shown potential towards various healthcare tasks~\cite{tu2025towards, singhal2023towards}. However, past works have also highlighted that LLMs struggle with issues such hallucinations, sociocultural biases~\cite{restrepo2024analyzing,liu2025culturevlm}, and misalignment with experts~\cite{chandra-etal-2025-lived}. Advancing model performance on healthcare related tasks requires both collecting richer data and deeper stakeholder involvement. While stakeholders have started collaborating with NLP researchers, access to high-quality data remains a challenge. Synthetic data generation offers a potential solution, especially with advancement in reasoning capabilities of LLMs. However, many question remain unanswered. How do we capture emerging or under-represented experiences when LLMs themselves are trained on existing web data? How can we ensure synthetic data mitigates bias and presents real-world lived experiences of humans? Addressing these questions is a pre-requisite for developing safer and more robust LLMs for healthcare domain. Holistically evaluating the efficacy of LLMs within high-risk domains remains an open challenge. While objective measures such as accuracy and win-rates~\cite{li2024mediq, liu2025exploring} remain popular metrics because of their objectivity and scalability, more subjective aspects of performance get overlooked. In this work, we evaluated conversations in a more holistic manner using clinically grounded metrics. However, subjective measures introduce their own limitations of generalizability. Hence, an open question remains, how can we develop evaluation metrics and frameworks that go beyond simplistic accuracy based metrics while maintaining the reliability,  scalability across diverse settings?

\noindent\textbf{Potential Risks}: While our work highlights the potential of using LLMs to simulate patient-sensemaker conversations, and some findings highlighted the capability of reasoning models towards conducting multi-turn mental health conversations, these findings should be interpreted with caution. While conversation generated through~\method framework encompass patients' goals, intents, and personality traits, but do not include other aspects of lived experiences, which might significantly affect the flow of the conversation.

\section{Limitations}
\label{sec:limitations}

While our work presents novel contribution on multiple fronts, it is important to acknowledge the limitations of our work. While~\method framework provides a method for generating realistic mental health sensemaking conversations, it is important to recognize both patient and sensemaker actors are modeled by LLMs which may not be able to capture the full range of emotions, variability and lived experiences of human patients and healthcare providers. Additionally, while grounding the conversations on the~\emph{6-function model}~\cite{de2009endpoints, king2013best} provides clinically sound structure to the conversation, the number of messages exchanged in each phase can vary by a great extent in real-world setting, something which is not fully captured in~\dataset~dataset. Finally, there could be other axes of evaluation along which one could evaluate multi-turn mental health conversation, something we may have missed in this work. Despite these challenges, our work provides a robust framework for generating synthetic clinical conversations grounded in real world settings, a large-scale conversational dataset, and a robust framework for assessing the capabilities of LLMs in multi-turn mental health conversations.

\section{Ethical Considerations}
\label{sec:ethical_considerations}

We used the clinical case-studies from~\citet{jin2021disease} which are publicly available for use under the MIT License. We further ensured that all data used was de-identified. As our study did not involve working with human participants, the Institutional Review Board (IRB) classified it as non-human subjects research, exempting it from IRB approval. We aim to provide the code and~\dataset~dataset publicly available on a GitHub repository upon acceptance of the paper.

Our study presents a systematic approach for generating synthetic data for meantal health conversations and evaluating LLM performance for such conversations. However, our framework is designed for research \& prototyping purposes and is not intended to replace licensed medical professionals. Using our framework without guardrails and proper oversight in real-world settings could pose risks to patient safety. Additionally, our findings should be complemented with thorough human evaluation to ensure the reliability and safety of models in real-world settings.

\section{Acknowledgments}

Chandra, Sriraman, Khanuja, and De Choudhury were partly supported through National Science Foundation grant 2230692, National Institute of Mental Health grants R01 MH117172 and P50 MH115838, and a grant from the American Foundation for Suicide Prevention. Further, this research project has benefitted from the Microsoft Accelerate Foundation Models Research (AFMR) grant program. The findings, interpretations, and conclusions of this paper are those of the authors and do not represent the official views of National Science Foundation, National Institute of Mental Health, American Foundation for Suicide Prevention or Microsoft.


\bibliography{custom}

\appendix

\begin{table*}[ht]
\centering
\begin{tabular}{p{2.8cm} p{4.2cm} p{7.2cm}}
\toprule
\textbf{Trait} & \textbf{Definition} & \textbf{Sensemaker-patient Conversational Style}\\
\midrule
Openness & Imagination, curiosity, comfort with novel, unusual, and abstract ideas. & \textbf{High}: Ask \emph{why}/\emph{how} questions about symptoms and medicines. Explore alternatives. Receptive to innovative therapies. \newline
\textbf{Low}: Prefer familiar terms and concrete instructions. May resist complex or less common therapies.\\
\midrule
Conscientiousness & Organization, reliability, self‑discipline, goal‑orientation. & \textbf{High}: Bring detailed symptom logs and timelines. Seek precise next steps. \newline
\textbf{Low}: Offer vague histories. Forget previous conversations. Prefer simplified plans.\\
\midrule
Extraversion & Sociability, talkativeness, interactions. & \textbf{High}: Give rich detail. Engage in small talk. Frequently ask questions. \newline
\textbf{Low}: Give succinct answers. Prefer structured questioning. Refrain from talking when necessary. \\
\midrule
Agreeableness & Cooperation, trust, empathy, prosocial orientation. & \textbf{High}: Readily agree and express understanding. May conceal doubts to avoid conflict. \newline
\textbf{Low}: Challenge recommendations. Negotiate treatment choices. Openly voice dissatisfaction or concerns.\\
\midrule
Neuroticism & Emotional reactivity, anxiety, mood instability. & \textbf{High}: Focus on worries. List multiple symptoms. Seek repeated reassurance and empathetic validation. \newline
\textbf{Low}: Remain calm and concise. Handle brisk, technical explanations without heightened distress.\\
\bottomrule
\end{tabular}
\caption{Definition of Big Five personality traits and typical sensemaker–patient conversation styles.}
\label{tab:appendix_big-5-examples}
\end{table*}

\section{Big Five Personality Traits}
\label{sec:appendix_big_5_personality}

Each patient persona is instantiated with specific personality traits that modulate their conversational behavior. For example, patients with high-\emph{Agreeableness} tend to seek social harmony. They use affiliative language (e.g. ``I totally understand your suggestions''), and readily consent to medical recommendations. However, they might under-report side‑effects or doubts to avoid ``bothering'' the doctor, which can obscure important clinical information. In contrast, low-\emph{Agreeableness} patients are more skeptical and assertive. They may challenge diagnoses, question procedures, and bargain over medications, requiring more negotiation from the physician. Similarly, High‑\emph{Neuroticism} often display heightened anxiety. They may focus on worst-case scenarios (e.g., ``Could this be cancer?''), describe multiple vague symptoms, and seek repeated reassurance. Such interactions tend to be emotionally charged and require empathetic listening and gradual explanation strategies. 
Conversely, low-\emph{Neuroticism} are more emotionally stable and pragmatic. They typically provide concise, factual accounts and are receptive to succinct, technical communication. Table~\ref{tab:appendix_big-5-examples}

\section{Health Literacy}
\label{sec:appendix_patient_health_literacy}
We define two health literacy levels building upon \citet{healthlit2009}, which used factors such as ability to describe diseases, self-reporting conditions and navigating healthcare systems to assess health literacy in Americans (as part of the National Assessment of Adult Literacy). For example, a patient with basic medical literacy responded to a sensemaker's diagnosis with "\textit{I guess I’d like to know more, but I don’t really understand how all this works. It’s just... ever since..}'', while one with advanced literacy generated "\textit{Yes, that sounds right. My father had similar symptoms, and they started when he was 33. Now I’m experiencing these involuntary movements, especially in my arms and hands .. genetic testing can confirm this.}''. Detailed characteristics for these levels are presented in Table~\ref{tab:patient_health_literacy}.

\begin{table*}[ht]
\centering
\begin{tabular}{p{4.2cm}|p{10.5cm}}
\toprule
\textbf{\small Health Literacy Level} & \textbf{\small Characteristics}\\
\midrule
\small Basic & \small > Struggles to understand basic medical terms and body parts. \newline
> Rarely describes symptoms beyond "pain" or "sick". \newline
> May use vague or informal terms instead of specific symptoms. \newline
> Often cannot follow written medical instructions. \newline
> May avoid seeking medical care due to communication barriers.\\
\midrule
\small Advanced & \small  > Can provide detailed symptom descriptions including onset and triggers. \newline
> Can describe subtle symptom variations and patterns. \newline
> Understands complex medical terminology. \newline
> Able to discuss medication effects and interactions. \newline
> Able to research and evaluate health information from reliable sources. \newline
> Maintains personal health records effectively.\\
\bottomrule
\end{tabular}
\caption{Definition of two health literacy levels used in the patient profiles.}
\label{tab:patient_health_literacy}
\end{table*}

\section{Patient Intentions}
\label{sec:appendix_patient_intentions}
While the filtered MedQA case studies involve a patient with some symptoms looking for a diagnosis, they are not tailored towards sensemaker conversations, which would start off with an overarching intent conveyed by the patient (either implicitly, or if explicitly asked by the sensemaker). The intent adds a unique vignette to help guide the conversation with the sensemaker at the early stages. To mimic a real-world setting, we refer to~\citet{aydin2024large}, which analyzed 201 works studying usage of AI in healthcare to characterize six themes of how patients use LLMs in these settings and ~\citet{lipersonalized2024}, which studied works on using AI to personalize healthcare. We filtered out themes which involve indirect use of LLMs in healthcare (e.g: generating patient education materials, optimizing doctor-patient interaction processes like medical consent forms) resulting in three themes: 1) \textit{Interpreting medical information from a patient perspective}, 2) \textit{Providing lifestyle recommendations and improving health literacy} and 3) \textit{Personalizing healthcare journeys}. For example, the starting message for a case study augmented with intention (1) is: "\textit{A few days ago, I was feeling really agitated and not myself—I don’t even remember much of what happened, but I was told I had to be sedated. I’m trying to make sense of all this and figure out what’s going on with me. Can you help?}''. Detailed definitions and examples of these intents are provided in Table \ref{tab:patient_intent_definitions}.

\begin{table*}[ht]
\centering
\begin{tabular}{p{2.8cm}|p{4.2cm}|p{7.2cm}}
\toprule
\textbf{\small Patient Intent} & \textbf{\small Definition} & \textbf{\small Sample Patterns}\\
\midrule
\small Interpreting medical information from a patient perspective & \small Patient aims to use LLMs to simplify complex medical terminology and concepts so that patients can more easily understand diagnoses, procedures, and general health information. & \small > Patient can provide their symptoms and past medical history to receive lay-person friendly explanations of their condition. \newline
\small > Patients can ask questions about their condition and receive clear, concise answers. \newline
\small > Patients can receive personalized advice on how to manage their health.\\
\midrule
\small Providing lifestyle recommendations and improving health literacy & \small Patient aims to use LLMs to seek lifestyle change recommendations and debunking myths about health and wellness. & 
\small > Patients can ask questions about lifestyle changes and receive recommendations. \newline
\small > Patients can query about the myths surrounding health and wellness. \\
\midrule
\small Personalizing \newline healthcare journeys & \small Patient aims to use LLMs to tailor educational content and recommendations based on individual patient data, resulting in more relevant and actionable advice. & \small > Patients enter personal health goals (like weight loss or improved mobility), asking LLMs for targeted tips. \newline
\small > Patients provide feedback on what they do or do not understand, and LLMs adjusts explanations accordingly. \\
\bottomrule
\end{tabular}
\caption{Definition and sample patterns of three high-level patient intents used to augment medical case studies.}
\label{tab:patient_intent_definitions}
\end{table*}

\section{Patient Goals}
\label{sec:appendix_patient_goals}
While intents serve as high-level motivations to initiate conversations with a sensemaker, patient goals are concrete outcomes they wish to achieve by the end of the conversation. Similar to intents, MedQA case studies do not express explicit patient goals as they are not conversational in nature. Again, to mimic a real-world setting, we use ~\citet{auriemma2024classification}, which carried out 338 "goals of care" conversations with 85 patients, resulting in three major themes: 1) \textit{Maximize comfort and avoid suffering}, 2) \textit{Maintain or improve cognitive or physical functioning} and 3) \textit{Extend longevity or survival}. Detailed definitions and examples of these intents are provided in Table \ref{tab:patient_goal_definitions}. Similar to the intents, we prompt GPT-4o to augment each patient profile with the single most appropriate goal given the medical facts of the case study (prompt in \ref{tab:appendix_patient_goal_augmentation_table}). "\textit{I appreciate you taking my feelings seriously. Right now, what would feel most helpful is anything that keeps me comfortable and minimizes my suffering. I don’t want to focus on things that might prolong my life if they make me feel worse in the process.}'' is an example of goal (1) manifesting explicitly in a patient message in response to a decision making statement from the sensemaker.


\begin{table*}[ht]
\centering
\begin{tabular}{p{4.2cm}|p{10.5cm}}
\toprule
\textbf{\small Patient Goal} & \textbf{\small Definition}\\
\midrule
\small Maximize comfort and \newline avoid suffering & \small Patient’s goal is to maximize comfort and avoid suffering. Includes seeking interventions to promote comfort (e.g., pain control) and avoiding interventions that would increase discomfort, even at the expense of decreasing longevity.\\
\midrule
\small Maintain or improve cognitive or physical functioning & \small Patient’s goal is to maintain or improve cognitive or physical functioning by undergoing medical care aimed at preventing or reversing dysfunction, even if that medical care would increase discomfort. However, care that would increase survival/longevity without preservation or improvement in function is generally avoided.\\
\midrule
\small Extending longevity or survival & \small Patient’s goal is to live as long as possible without limitations on care. Extending longevity or survival is prioritized over maximizing function or comfort.\\
\bottomrule
\end{tabular}
\caption{Definition of three concrete patient goals used to augment medical case studies.}
\label{tab:patient_goal_definitions}
\end{table*}

\begin{table*}[ht]
\centering
\small
\begin{tabular}{p{0.3\textwidth}|p{0.6\textwidth}}
\toprule
\textbf{Stage} & \textbf{Example Conversation} \\
\midrule
Fostering the Relationship & Hello, how can I help you today? \\
\midrule
Gathering Information & 
\begin{tabular}[t]{@{}l@{}}
Can you tell me more about your concerns? \\
Can you tell me more about your personal and family history? \\
What are your expectations from me? \\
Can you expand on this point? \\
Based on your description, it sounds like you're feeling... \\
I understand you're feeling... \\
I'm hearing that...
\end{tabular} \\
\midrule
Providing Information & 
Your test results indicate that your cholesterol levels are high. This means you are at a greater risk for heart disease, but we can work on strategies like diet and exercise to manage it effectively. \\
\midrule
Decision Making & 
There are a few different treatment options available. Would you like me to walk you through them so we can decide together which one aligns best with your lifestyle and preferences? \\
\midrule
Enabling Disease and Treatment-Related Behavior & 
Managing diabetes can be overwhelming, but breaking it down into small steps—like checking your blood sugar daily—can make it easier. Would you like me to recommend some support groups or resources? \\
\midrule
Responding to Emotions & 
I can see that this diagnosis is really affecting you. It’s completely understandable to feel this way. I'm here to support you, and we can discuss ways to help manage both your symptoms and the emotional impact. \\
\bottomrule
\end{tabular}
\caption{
The \emph{6-function model} of medical communication~\citep{de2009endpoints, king2013best}, which characterizes key communicative functions in clinical interactions, accompanied by representative conversation for each stage.
}
\label{tab:example_conversation}
\end{table*}

\section{Sensemaker Next Stage Determination Sub-module Prompt}
\label{sec:appendix_sensemaker_next_stage_determination_prompt}
\begin{table*}[ht]
    \centering
    \small
    \begin{tabular}{p{0.10\linewidth} | p{0.85\linewidth}}
    \toprule
    \textbf{Type} & \textbf{Prompt}\\
    \toprule
         \textbf{System Prompt}
            &Given the current patient message (PATIENT\_MESSAGE) a list of facts about the patient (PATIENT\_FACTS) and a list of statements previously made by you (YOUR\_MEMEORY). Determine whether the conversation should remain in the current stage (Fostering the Relationship) or transition to the next stage (Gathering Information).If you decide to stay in the current stage then your OUTPUT\_STAGE should be 'STAYCURRENTSTAGE' and if you determine to transition to the next stage then your output should be 'MOVENEXTSTAGE'.\\
            &\\
            &Return Format:\\
            &OUTPUT\_REASONING: $<$your reasoning for the stage determination$>$\\
            &OUTPUT\_CONFIDENCE: $<$your confidence score to transition to the next stage$>$\\
            &OUTPUT\_STAGE: $<$MOVENEXTSTAGE or STAYCURRENTSTAGE$>$\\
            &\\
            &Guidelines:\\
            &- Refer to POTENTIAL\_NEXT\_STAGE\_REASONING and then reach your conclusion.\\
            &- Before determining if you should stay in the same stage or transition to the next stage, you should generate a step-by-step reasoning (OUTPUT\_REASONING) for your conclusion.\\
            &- You should also generate a score between 1 and 10 (OUTPUT\_CONFIDENCE) indicating your confidence in transitioning to the next stage, if the score is between 1 and 3, you should definately stay in the current stage, if the score is between 4 and 6, then you should stay in the stage for one or two more turns, and if the score is 7 or above then you should move to the next stage.\\
            &- The conversation should remain in the current stage if the user has not yet fully engaged with or completed its objectives.\\
            &- The conversation should move to the next stage only when the current stage has been meaningfully completed, ensuring a natural transition.\\
            &\\
            &Warnings:\\
            &- Do not keep the conversation stuck in the same stage for multiple iterations unless necessary. If progression is unclear, consider whether the user is engaging sufficiently before deciding.\\
         &\\
         \midrule
         \textbf{User Prompt}
         &POTENTIAL\_NEXT\_STAGE\_REASONING: <next\_stage\_reasoning>\\
         &PATIENT\_MESSAGE: <current patient message>\\
         &PATIENT\_FACTS: <atomic facts about the patient till that timestep>\\
         &YOUR\_MEMORY: <sensemaker messages till that timestep>\\
         \bottomrule
    \end{tabular}
    \caption{(Sensemaker Module) Prompt used for the Gathering Information stage determination.}
    \label{tab:appendix_next_stage_determination_prompt_table}
\end{table*}
Table~\ref{tab:appendix_next_stage_determination_prompt_table} provides the prompt for the next stage determination module and Table~\ref{tab:appendix_next_stage_determination_dictionary} provides the rationale for stage movements provided in the prompt.\\ 
\begin{table*}[ht]
    \centering
    \small
    \begin{tabular}{p{0.10\linewidth} | p{0.85\linewidth}}
    \toprule
    \textbf{Current Stage} & \textbf{Next Stage Reasoning}\\
    \toprule
        Fostering the Relationship & The next stage would be 'Gathering Information' because an initial relationship between you and the patient has been established. Move on when the 'Fostering the Relationship' stage has provided a welcoming space and the patient starts to openly describe their concerns, feelings, or challenges. If they seem hesitant or reserved, stay in 'Fostering the Relationship' longer to encourage sharing. \\
        &\\
        Gathering Information & The next stage would be 'Providing Information' only when you have received enough information from the patient about their current condition and symptoms for you to make a diagnosis. If you feel you need more time to gather information to make a confident diagnosis, stay in 'Gathering Information' longer. \\
        &\\
        Providing Information & You should move to the next stage which is 'Decision Making', NO MATTER WHAT. \\
        &\\
        Decision Making & The next stage would be 'Responding to Emotions' because the patient has understood the lifestyle and non-clinical suggestions made by you to alleviate their current condition. Move forward from the 'Decision Making' stage when the patient has acknowledged your suggestions. If their responses suggest they still need more clarity or direction, stay in 'Decision Making' to provide additional support.\\
        &\\
        Responding to Emotions & The next stage would be 'exit' because the conversation has reached its end. Move forward from the 'Responding to Emotions' when you validated their emotions with an empathetic response and the patient has replied with an affirmative message. Do not focus on coping mechanisms in this stage.\\
        \bottomrule
    \end{tabular}
    \caption{Reasoning provided to the prompt in Table~\ref{tab:appendix_next_stage_determination_prompt_table}.}
    \label{tab:appendix_next_stage_determination_dictionary}
\end{table*}

\section{Sensemaker Fact Extraction and Matching Sub-module Prompt}
\label{sec:appendix_sensemaker_fact_extraction_and_matching_prompt}

We use the prompt in Table~\ref{tab:appendix_fact_extraction_prompt_table} to extract atomic facts from the patient's current message and then use the prompt in Table~\ref{tab:appendix_fact_matching_prompt_table} to find if there are any matching facts already present in the memory.
\begin{table*}[ht]
    \centering
    \small
    \begin{tabular}{p{0.10\linewidth} | p{0.85\linewidth}}
    \toprule
    \textbf{Type} & \textbf{Prompt}\\
    \toprule
         \textbf{System Prompt}
         &You are given a message provided by the patient as USER\_MESSAGE, your task is to extract explicitly stated atomic facts about the patient from the message. Here, an atomic fact is defined as new information provided by the patient which can increase your holistic understanding of the patient's condition. Each atomic fact should carry an entirely different piece of explicitly stated fact, and should be independent of other atomic facts in the list.\\
        &\\
        &Your output should strictly be a list of atomic facts, with each item starting with "\# ". Do not include other formatting. In addition, each of the atomic facts in the list should be in third person narration.\\
        &\\
        &Keep in mind that each atomic fact is different from other atomic facts in the list. Do not add any new fact which was not present in the USER\_MESSAGE.\\
        &\\
         \midrule
         \textbf{User Prompt}
         &USER\_MESSAGE: <current patient message>\\
         &ATOMIC\_FACTS:\\
         \bottomrule
    \end{tabular}
    \caption{(Sensemaker Module) Prompt used for Fact Extraction.}
    \label{tab:appendix_fact_extraction_prompt_table}
\end{table*}

\begin{table*}[ht]
    \centering
    \small
    \begin{tabular}{p{0.10\linewidth} | p{0.85\linewidth}}
    \toprule
    \textbf{Type} & \textbf{Prompt}\\
    \toprule
         \textbf{System Prompt}
            &You are an expert fact matching agent. You are given a list of NEW\_FACTS about a patient, and a list of already present facts in MEMORY\_OF\_PATIENT\_FACTS. Your task is to analyze if facts listed in NEW\_FACTS list are 'FactPresent' or 'FactNotPresent' in MEMORY\_OF\_PATIENT\_FACTS. For this task you take each fact from NEW\_FACTS list one by one and check if it is present in MEMORY\_OF\_PATIENT\_FACTS or not. For this checking you should look at the facts in terms of their meaning and context and not the exact words. The classification for each NEW FACT should be independent of other NEW\_FACTS.\\
            &\\
            &Your output should strictly follow the format: \# <fact> : <'FactNotPresent' or 'FactPresent'>. Where fact presents one fact from NEW\_FACTS. Output the labels for each fact in NEW\_FACTS and keep the fact text as it is (do not change the words).\\
            &\\
            &You should not change the words of the fact, only add the label (either 'FactPresent' or 'FactNotPresent'). Your output should have one fact and its label per line.\\
            &\\
         \midrule
         \textbf{User Prompt}
         &NEW\_FACTS: <facts extracted from the current patient message>\\
         &MEMORY\_OF\_PATIENT\_FACTS: <atomic facts about the patient till that timestep>\\
         &ANSWER: \\
         \bottomrule
    \end{tabular}
    \caption{(Sensemaker Module) Prompt used for Fact Matching.}
    \label{tab:appendix_fact_matching_prompt_table}
\end{table*}

\section{Sensemaker Message Generation Sub-module Prompts}
\label{sec:appendix_sensemaker_message_generation_prompts}

For the message generation sub-module, we have a similar structure for all the stages as shown in Table ~\ref{tab:appendix_all_stages_message_generation_prompt_table} with slight modifications in the gathering information and providing information stage as shown in Table~\ref{tab:appendix_gathering_information_message_generation_prompt_table} and Table~\ref{tab:appendix_providing_information_message_generation_prompt_table} system prompts, respectively.
\begin{table*}[ht]
    \centering
    \small
    \begin{tabular}{p{0.10\linewidth} | p{0.85\linewidth}}
    \toprule
    \textbf{Type} & \textbf{Prompt}\\
    \toprule
         \textbf{System Prompt}
            &You are an expert sensemaker who helps the patients to make sense of their current clinical situation. Keeping the conversation's current stage goal and example messages in mind (given by CURRENT\_STAGE\_GOAL and CURRENT\_STAGE\_EXAMPLES), list of facts already known about the patient (PATIENT\_FACTS), and the list of statements already made by you (YOUR\_MEMORY), and the user's current message (USER\_MESSAGE), your goal is to generate three plausible and logical messages to be said to the patient that you have not said yet. You should first generate a reasoning for yourself and then generate the three messages. Your messages can be a question or a statement.\\
            & \\
            &Your output should strictly be in the following format:\\ &OUTPUT\_REASONING: <your step-by-step reasoning>\\
            &OUTPUT\_MESSAGE: <3 plausible messages to the patient each on a new line and starting with "\# ">. Do not include other formatting.\\
            & \\
            &All the three plausible messages should be different from each other. If you are asking a question in a message, then only ask one question at a time in that message. Your message should be concise and to the point.\\
            & \\
            &YOUR\_MEMORY: <sensemaker's messages till that timestep>\\
            & \\
            &PATIENT\_FACTS: <atomic facts about the patient till that timestep>\\
            & \\
            &CURRENT\_STAGE\_GOAL: <goal that needs to be achieved in the current stage>\\
            &\\
            &CURRENT\_STAGE\_EXAMPLES: <examples of messages for the current stage>\\
            &\\
         \midrule
         \textbf{User Prompt}
         &USER\_MESSAGE: <current patient message>\\
         \bottomrule
    \end{tabular}
    \caption{(Sensemaker Module) Prompt used for message generation in the all other stages.}
    \label{tab:appendix_all_stages_message_generation_prompt_table}
\end{table*}

\begin{table*}[ht]
    \centering
    \small
    \begin{tabular}{p{0.10\linewidth} | p{0.85\linewidth}}
    \toprule
    \textbf{Type} & \textbf{Prompt}\\
    \toprule
         \textbf{System Prompt}
            &You are an expert sensemaker who helps the patients to make sense of their current clinical situation. Keeping the conversation's current stage goal and example messages in mind (given by CURRENT\_STAGE\_GOAL and CURRENT\_STAGE\_EXAMPLES), list of facts already known about the patient (PATIENT\_FACTS), and the list of statements already made by you (YOUR\_MEMORY), your goal is to generate three plausible and logical messages to be said to the patient that you have not said yet. Before generating the message you should generate a step-by-step reasoning taking into account the facts you already know about the patient. In your reasoning you should think about the possible diagnosis hypotheses, and then generate the three messages for the patient that helps in gathering more information to either confirm or reject the diagnosis hypotheses. \\
            & \\
            &Your output should strictly be in the following format:\\ &OUTPUT\_REASONING: <your step-by-step reasoning>\\
            &OUTPUT\_MESSAGE: <3 plausible messages to the patient each on a new line and starting with "\# ">. Do not include other formatting.\\
            & \\
            &All the three plausible messages should be different from each other. If you are asking a question in a message, then only ask one question at a time in that message. Your message should be concise and to the point.\\
            & \\
            &YOUR\_MEMORY: <sensemaker's messages till that timestep>\\
            & \\
            &PATIENT\_FACTS: <atomic facts about the patient till that timestep>\\
            & \\
            &CURRENT\_STAGE\_GOAL: <goal that needs to be achieved in the current stage>\\
            &\\
            &CURRENT\_STAGE\_EXAMPLES: <examples of messages for the current stage>\\
            &\\
         \midrule
         \textbf{User Prompt}
         &USER\_MESSAGE: <current patient message>\\
         \bottomrule
    \end{tabular}
    \caption{(Sensemaker Module) Prompt used for message generation in the Gathering Information stage.}
    \label{tab:appendix_gathering_information_message_generation_prompt_table}
\end{table*}

\begin{table*}[ht]
    \centering
    \small
    \begin{tabular}{p{0.10\linewidth} | p{0.85\linewidth}}
    \toprule
    \textbf{Type} & \textbf{Prompt}\\
    \toprule
         \textbf{System Prompt}
            &You are an expert sensemaker who helps the patients to make sense of their current clinical situation. Keeping the conversation's current stage goal and example messages in mind (given by CURRENT\_STAGE\_GOAL and CURRENT\_STAGE\_EXAMPLES), list of facts already known about the patient (PATIENT\_FACTS), and the list of statements already made by you (YOUR\_MEMORY), your goal is to provide the patient with the diagnosis and its explanation to the patient. Before generating the diagnosis and explanation message you should generate a step-by-step reasoning for yourself taking into account the facts you already know about the patient. \\
            & \\ 
            &Your output should strictly be in the following format:\\ &OUTPUT\_REASONING: <your step-by-step reasoning>\\
            &OUTPUT\_MESSAGE: <3 plausible messages to the patient each on a new line and starting with "\# ">. Do not include other formatting.\\
            & \\
            &You should always provide a diagnosis and if you cannot find a diagnosis your message should strictly be "I apologize but I am unable to diagnose you at the moment."\\
            & \\
            &All the three plausible messages should be different from each other. If you are asking a question in a message, then only ask one question at a time in that message. Your message should be concise and to the point.\\
            & \\
            &YOUR\_MEMORY: <sensemaker's messages till that timestep>\\
            & \\
            &PATIENT\_FACTS: <atomic facts about the patient till that timestep>\\
            & \\
            &CURRENT\_STAGE\_GOAL: <goal that needs to be achieved in the current stage>\\
            &\\
            &CURRENT\_STAGE\_EXAMPLES: <examples of messages for the current stage>\\
            &\\
         \midrule
         \textbf{User Prompt}
         &USER\_MESSAGE: <current patient message>\\
         \bottomrule
    \end{tabular}
    \caption{(Sensemaker Module) Prompt used for message generation in the Providing Information stage.}
    \label{tab:appendix_providing_information_message_generation_prompt_table}
\end{table*}

Table~\ref{tab:appendix_stage_goal_dictionary} shows the goals associated with each state that are passed to the system prompt as CURRENT\_STAGE\_GOAL.

\begin{table*}[ht]
    \centering
    \small
    \begin{tabular}{p{0.10\linewidth} | p{0.85\linewidth}}
    \toprule
    \textbf{Current Stage} & \textbf{Goal}\\
    \toprule
        Fostering the Relationship & Your goal is to build a trusting, open, and collaborative relationship with the patient by demonstrating empathy, honesty, and respect. You need to create a safe and supportive environment where the patient feels heard, valued, and comfortable sharing sensitive information. For this you can use the following strategies:\\
        &1. Invite the patient to share their story in their own words.\\
        &2. Respond with empathy and compassion to patient's concerns.\\
        &3. Express compassion and commitment. Let the patient know that you care about their well-being and are committed to helping them.\\
        &4. If the patient is not comfortable sharing their story, you can ask them to share their concerns and you can respond with empathy and compassion. \\
        &\\
        Gathering Information & Your goal is to develop a comprehensive understanding of the patient’s needs, concerns, and medical history by exploring their condition from both biological and psychosocial perspectives. This understanding will allow you to support the patient in achieving their goals and expectations for the conversation. For this you can use the following strategies:\\
        &1. Ask open ended questions related to patient's concerns to gather information about patients current state, their personal and family history.\\
        &2. Listen actively and ask follow-up questions to understand the situation better.\\
        &3. Elicit patient's perspective of the problems and their expectations from you.\\
        &4. Clarify and summarize the information gathered from the patient to ensure understanding.\\
        &\\
        Providing Information & At this stage you want to provide a potential diagnosis to the patient for their concerns. You should not ask any questions in this stage and rather provide a potential diagnosis to patient based on the their personal history, family history, concerns and other details. \\
        &\\
        Decision Making & "At this stage you should address any medical queries posed by the patient regarding your diagnosis and suggest only lifestyle or non-clinical changes to the patient to alleviate their illness based on their diagnosis. You should make sure that the changes suggested by you are based on the patient's preferences and your all previous knowledge about them. For this you can use the following strategies:\\
        &1. Ask for patient's preferences and suggestions regarding the lifestyle changes or other non-clinical changes.\\
        &2. Suggest lifestyle changes based on the patient's preferences and your all previous knowledge about them.\\
        &3. If the patient is not comfortable with the suggestions, you can ask them to suggest their own lifestyle or non-clinical changes.\\
        &\\
        Responding to Emotions & You need to recognize and address any emotional aspect of the illness by offering empathay, reassurance, and psychological support in your messages. For this you can use the following strategies:\\
        &1. Offer empathay and reassurance to the patient.\\
        &2. Listen to the patient's concerns and offer psychological support.\\
        &3. If the patient is not comfortable sharing their concerns, you can ask them to share their emotions and you can respond with empathy and support.\\
        &\\
        Exit & The conversation has reached its end and we need to conclude the conversation.\\
\bottomrule
    \end{tabular}
    \caption{Current stage goal provided to the sensemaker message generation prompt}
    \label{tab:appendix_stage_goal_dictionary}
\end{table*}

\section{Sensemaker Message Redundancy Checking Sub-module Prompt}
\label{sec:appendix_sensemaker_redundancy_checking_prompt}

We use the prompt in Table~\ref{tab:appendix_message_redundancy_prompt_table} to check if the candidate messages generated by the sensemaker are redundant or not.
\begin{table*}[ht]
    \centering
    \small
    \begin{tabular}{p{0.10\linewidth} | p{0.85\linewidth}}
    \toprule
    \textbf{Type} & \textbf{Prompt}\\
    \toprule
         \textbf{System Prompt}
            &You are an expert sensemaker who helps the patients to make sense of their current clinical situation. Given a list of statements already made by you (STATEMENT\_MEMORY), and a list of candidate statements (CANDIDATE\_STATEMENTS), your task is to identify if there are any redundant statements in the CANDIDATE\_STATEMENTS.\\
            & \\
            &Your output should strictly follow the format: \# <statement> : <'RedundantStatement' or 'RedundantNotStatement'>. Where statement presents one statement from CANDIDATE\_STATEMENTS. Output the labels for each statement in CANDIDATE\_STATEMENTS and keep the statement text as it is (do not change the words) and always start the line with '\#'.\\
            & \\
            &You should not change the words of the statement, only add the label. Your output should have one statement and its label per line. Below are some additional instructions:\\
            & \\
            &1. You should not add any new statement which was not present in the STATEMENT\_MEMORY.\\
            & \\
            &2. If a new statement in the CANDIDATE\_STATEMENTS has a different phrasing but serves a similar context to any of the statements present in the STATEMENT\_MEMORY, it should be considered 'RedundantStatement'.\\
            \\
            &3. Classification for each statement in CANDIDATE\_STATEMENTS should be independent of other statements in the CANDIDATE\_STATEMENTS.\\
            &\\
         \midrule
         \textbf{User Prompt}
         &CANDIDATE\_STATEMENTS: <candidate statements that can be made in this particular stage>\\
         &STATEMENT\_MEMORY: <statements already made by the sensemaker>\\
         \bottomrule
    \end{tabular}
    \caption{(Sensemaker Module) Prompt used for checking message redundancy.}
    \label{tab:appendix_message_redundancy_prompt_table}
\end{table*}

\begin{table*}[ht]
    \centering
    \small
    \begin{tabular}{p{0.10\linewidth} | p{0.85\linewidth}}
    \toprule
    \textbf{Type} & \textbf{Prompt}\\
    \toprule
         \textbf{System Prompt}
&You are an assistant that is given information about a patient.\\
&Break the following patient information into a list of independent atomic facts, with one piece of information in each statement. Each fact should only include the smallest unit of information, but should be self-contained.\\
&\\
&Criteria:\\
&1. Only extract medical facts that would be relevant in a conversation with a health-care provider.\\
&2. First, list out facts related to a patient's vitals (ONLY temperature, blood pressure, pulse, respiratory rate) under "PATIENT VITAL FACTS:".\\
&3. If there are no PATIENT VITAL FACTS just leave the section empty. \\
&4. List out the remaining medical atomic facts under the section "PATIENT FACTS:".\\
&\\
         \midrule
         \textbf{User Prompt}
&Patient Information:\\
&<patient\_data>\\
&\\
&Respond with the list of atomic facts for each section "PATIENT VITAL FACTS:" and "PATIENT FACTS:" (no formatting) and nothing else, prepend each fact with a '- '. No sub-lists are allowed.\\
         \bottomrule
    \end{tabular}
    \caption{(Patient Module) Prompt used for extracting patient facts.}
    \label{tab:appendix_patient_fact_extraction_prompt_table}
\end{table*}

\begin{table*}[!htbp]
    \centering
    \small
    \begin{tabular}{p{0.10\linewidth} | p{0.85\linewidth}}
    \toprule
    \textbf{Type} & \textbf{Prompt}\\
    \toprule
         \textbf{System Prompt}
&You are an expert psychiatrist/psychologist/psychotherapist. Given a patient case study, your task is to pick the most appropriate goal for that patient's consultation with an LLM. You are given a case study (provided as CASE\_STUDY) and a list of patient conversation goal vignettes in the format (VIGNETTE\_NAME, VIGNETTE\_DEFINITION, VIGNETTE\_PATTERNS). You need to pick the most appropriate vignette for the given case study.\\
&\\
&Think step by step and first provide your rationale (under "RATIONALE:") in around 30 words and then at the last output the VIGNETTE\_NAME for the most logical vignette for the given case study.\\ 
&\\
&1. VIGNETTE\_NAME:\\
&VIGNETTE\_DEFINITION:\\
&\\
&2. VIGNETTE\_NAME:\\
&...\\
&\\
&Think step by step and first provide your rationale (under "RATIONALE:") and then at the last output the VIGNETTE\_NAME for the most logical vignette for the given case study. \\
         \midrule
         \textbf{User Prompt}
&CASE\_STUDY: <case\_study>\\
&Strictly follow the format: <ASSIGNED\_VIGNETTE\_NAME: VIGNETTE\_NAME>. Use the exact vignette name and nothing else.\\
         \bottomrule
    \end{tabular}
    \caption{(Patient Module) Prompt used to attach a concrete conversational goal to a case study. "Vignette" here refers to the goal.}
    \label{tab:appendix_patient_goal_augmentation_table}
\end{table*}

\begin{table*}[ht]
    \centering
    \small
    \begin{tabular}{p{0.10\linewidth} | p{0.85\linewidth}}
    \toprule
    \textbf{Type} & \textbf{Prompt}\\
    \toprule
         \textbf{System Prompt}
&You are an expert psychiatrist/psychologist/psychotherapist. Given a patient case study, your task is to pick the most appropriate motivation for that patient to consult an LLM. You are given a case study (provided as CASE\_STUDY) and a list of patient conversation intent vignettes in the format (VIGNETTE\_NAME, VIGNETTE\_DEFINITION, VIGNETTE\_PATTERNS). You need to pick the most appropriate vignette for the given case study.\\
&\\
&1. VIGNETTE\_NAME:\\
&VIGNETTE\_DEFINITION:\\
&VIGNETTE\_PATTERNS:\\
&\\
&2. VIGNETTE\_NAME:\\
&...\\
&\\
&Think step by step and first provide your rationale (under "RATIONALE:") and then at the last output the VIGNETTE\_NAME for the most logical vignette for the given case study. \\
         \midrule
         \textbf{User Prompt}
&CASE\_STUDY: <case\_study>\\
&Strictly follow the format: <ASSIGNED\_VIGNETTE\_NAME: VIGNETTE\_NAME>. Use the exact vignette name and nothing else.\\
         \bottomrule
    \end{tabular}
    \caption{(Patient Module) Prompt used to attach a high-level conversational intent to a case study. "Vignette" here refers to the intent.}
    \label{tab:appendix_patient_intent_augmentation_table}
\end{table*}

\begin{table*}[ht]
    \centering
    \small
    \begin{tabular}{p{0.10\linewidth} | p{0.85\linewidth}}
    \toprule
    \textbf{Type} & \textbf{Prompt}\\
    \toprule
         \textbf{System Prompt}
&You are an intelligent agent who is given a QUESTION from a doctor to a patient, and a LIST OF MEDICAL FACTS about the patient. Your task is to choose the facts which best answer the question, or respond "NO MATCH".\\
&\\
&Instructions:\\
&1. Given a LIST OF MEDICAL FACTS about a patient, choose a MAXIMUM of THREE facts from the list that when combined best answers the QUESTION. \\
&2. If NO fact matches the question, simply respond "NO MATCH" under "FACTS:".\\
&3. If facts are chosen, output them as a list, preserving the numbering from the original list. \\
&4. First provide a short reasoning under "REASONING:" before listing the facts under "FACTS:".\\
&5. Do not include any other formatting or extra information beyond the REASONING and given FACTS.\\
&\\
         \midrule
         \textbf{User Prompt}
&QUESTION: <question>\\
&LIST OF MEDICAL FACTS:\\
&<patient\_info>\\
&\\

&SOLUTION:\\
         \bottomrule
    \end{tabular}
    \caption{(Patient Module) Prompt used for selecting relevant medical facts to answer a question.}
    \label{tab:appendix_patient_answer_fact_selection_prompt_table}
\end{table*}
 
\begin{table*}[ht]
    \centering
    \small
    \begin{tabular}{p{0.10\linewidth} | p{0.85\linewidth}}
    \toprule
    \textbf{Type} & \textbf{Prompt}\\
    \toprule
         \textbf{System Prompt}
&You are an intelligent agent who is given a LIST OF MEDICAL FACTS about a patient, who is going to have a conversation with an AI healthcare provider. Your task is to construct a start message with the most important facts in first person, as if you are simulating the patient's persona and literacy while talking to the AI.\\
&\\
&To aid in this process, Big 5 personality traits are defined as:\\
&\\
&<personality\_trait\_definitions>\\
&\\
&Instructions:\\
&1. Given a LIST OF MEDICAL FACTS about a patient, choose a MAXIMUM of TWO relevant facts from the list to construct a starting message as a patient to an AI healthcare provider.\\
&2. The start message should be in first person, using the emotions, tone, word choice and intensity of a patient who has the levels of the Big 5 Personality Traits (on a 2-point Low/High scale) and Medical Literacy Level shown below.\\
&3. ONLY use the medical literacy level and personality definitions as a guide, DO NOT EXPLICITLY integrate any information from it into the answer.\\
&4. First provide a short reasoning under "REASONING:" before writing the start message under "START\_MESSAGE:".\\
&5. Do not include quotes, any other formatting, extra information beyond the REASONING and the START\_MESSAGE.\\
&\\
         \midrule
         \textbf{User Prompt}
&LIST OF MEDICAL FACTS:\\
&<patient\_info>\\
&\\
&Big 5 Personality Traits:\\
&1. Openness: <Openness\_score>\\
&2. Conscientiousness: <Conscientiousness\_score>\\
&3. Extraversion: <Extraversion\_score>\\
&4. Agreeableness: <Agreeableness\_score>\\
&5. Neuroticism: <Neuroticism\_score>\\
&\\
&MEDICAL LITERACY LEVEL:\\
&<medical\_literacy\_level>\\
&\\
         \bottomrule
    \end{tabular}
    \caption{(Patient Module) Prompt used for patient to start the conversation with the sensemaker.}
    \label{tab:appendix_patient_start_message_generation_table}
\end{table*}

\begin{table*}[!htbp]
    \centering
    \small
    \begin{tabular}{p{0.10\linewidth} | p{0.85\linewidth}}
    \toprule
    \textbf{Type} & \textbf{Prompt}\\
    \toprule
         \textbf{System Prompt}
&You are a truthful assistant that understands a patient's medical information, and you are trying to answer questions from an AI about the patient in first person, as if you are simulating the patient's persona and literacy while talking to the AI.\\
&\\
&To aid in this process, Big 5 personality traits are defined as:\\
&\\
&<personality\_trait\_definitions>\\
&\\
&Instructions:\\
&1. Using the list of medical facts given below, generate a conversational human-like response to an LLM in first person, using the emotions, tone, word choice and intensity of a patient who has the levels of the Big 5 Personality Traits (on a 2-point Low/High scale) and Medical Literacy Level shown below.\\
&2. Any information asked in the question that is NOT explicitly present in the facts can be assumed to be FALSE.\\
&3. If the question is just a statement diagnosing your condition, agree to it.\\
&4. ONLY use the medical literacy level and personality definitions as a guide, DO NOT EXPLICITLY integrate any information from it into the answer.\\
&5. DO NOT add ANY information in your answer that is NOT EXPLICITLY stated in the given facts.\\
&6. DO NOT suggest any treatment method yourself in the answer. \\
&7. ONLY respond with the answer, do NOT add any other formatting.\\
&\\
         \midrule
         \textbf{User Prompt}
&LIST OF MEDICAL FACTS:\\
&<patient\_info>\\
&\\
&QUESTION:\\
&<question>\\
&\\
&Big 5 Personality Traits:\\
&1. Openness: <Openness\_score>\\
&2. Conscientiousness: <Conscientiousness\_score>\\
&3. Extraversion: <Extraversion\_score>\\
&4. Agreeableness: <Agreeableness\_score>\\
&5. Neuroticism: <Neuroticism\_score>\\
&\\
&MEDICAL LITERACY LEVEL:\\
&<medical\_literacy\_level>\\
&\\
         \bottomrule
    \end{tabular}
    \caption{(Patient Module) Prompt used to generate answers given a non-zero number of selected facts. Note that in stages beyond Decision Making, point 2 in the instructions (closed-world assumption) is removed to aid natural conversation flow.}
    \label{tab:appendix_patient_answer_generation_table}
\end{table*}

\begin{table*}[!htbp]
    \centering
    \small
    \begin{tabular}{p{0.10\linewidth} | p{0.85\linewidth}}
    \toprule
    \textbf{Type} & \textbf{Prompt}\\
    \toprule
         \textbf{System Prompt}
&You are a truthful assistant that understands a patient's medical information, and you are trying to answer questions from an AI about the patient in first person, as if you are the patient with a specific persona and literacy talking to an AI.\\
&\\
&To aid in this process, Big 5 personality traits are defined as:\\
&\\
&<personality\_trait\_definitions>\\
&\\
&Instructions:\\
&1. Generate a conversational human-like response to an LLM in first person, using the emotions, tone, word choice and intensity of a patient who has the levels of the Big 5 Personality Traits (on a 2-point Low/High scale) and Medical Literacy Level shown below.\\
&2. If this question requests information about the patient's medical history, personal experience, medication, background or condition, respond in the negative stating "No ..." or "I'm not sure about ..." under "ANSWER:".\\
&3. If it is a question about decision-making, emotions, diagnosis or taking next steps, generate a conversational human-like response to an LLM in first person, responding in the affirmative ("Yes ...") as a patient.\\
&4. ONLY use the medical literacy level and personality definitions as a guide, DO NOT EXPLICITLY integrate any information from it into the answer.\\
&5. DO NOT add ANY NEW information to the answer. If asked about any such additional information, respond with "I'm not sure about ..." to them.\\
&6. First provide a short reasoning under "REASONING:" before writing the answer under "ANSWER:". Do not add any other formatting.\\
&\\
&Here are some examples to help you\\
&\\
&EXAMPLE 1:\\
&QUESTION: Have you tried any meditation-style exercises?\\
&REASONING: Exercises and lifestyle relates to the patient's personal information.\\
&ANSWER: No, I have not tried any meditation-related exercises.\\
&\\
&EXAMPLE 2: \\
&QUESTION: Would you like to explore some coping strategies related to your condition?\\
&REASONING: This is a question about next steps, hence the answer is in the affirmative.\\
&ANSWER: Yes, I would like to learn about some coping strategies.\\
&\\
&EXAMPLE 3: \\
&QUESTION: Looks like you are experiencing insomnia. When did you first start feeling these symptoms?\\
&REASONING: Though there is a diagnosis, the question part asks some medical history information.\\
&ANSWER: I'm not sure about when I first started feeling these symptoms.\\
&\\
&EXAMPLE 4: \\
&QUESTION: Looks like you are experiencing insomnia. What do you think about trying a medication routine to help with this?\\
&REASONING: There is a diagnosis and the question asks about next steps or advice, hence the answer is in the affirmative.\\
&ANSWER: I see, yes, I would like to learn about medications to help my situation.\\
&\\
         \midrule
         \textbf{User Prompt}
&LIST OF MEDICAL FACTS:\\
&<patient\_info>\\
&\\
&Big 5 Personality Traits:\\
&1. Openness: <Openness\_score>\\
&2. Conscientiousness: <Conscientiousness\_score>\\
&3. Extraversion: <Extraversion\_score>\\
&4. Agreeableness: <Agreeableness\_score>\\
&5. Neuroticism: <Neuroticism\_score>\\
&\\
&MEDICAL LITERACY LEVEL:\\
&<medical\_literacy\_level>\\
&\\
&QUESTION:\\
&<question>\\
&\\
         \bottomrule
    \end{tabular}
    \caption{(Patient Module) Prompt used to generate answers when there are no selected facts matching the question.}
    \label{tab:appendix_patient_unknown_answer_generation_table}
\end{table*}

\section{Qualitative Analysis}
\label{sec:appendix_qualitative_analysis}
We conducted a qualitative analysis for the lowest-scoring conversations across all three patient-centeric communication metrics, focusing on DeepSeek R1 as the sensemaker actor LLM
While the sensemaker stays empathetic and supportive, a lack of personalization emerges as a major theme across all metrics. However, this drawback means different things for each metric.\\
Sensemaker's dialogues such as \textit{"The combination of sleeplessness without tiredness and auditory hallucinations suggests your brain chemistry might be imbalanced, potentially indicating conditions like bipolar disorder or schizophrenia that need professional diagnosis."} and \textit{"Your symptoms suggest moderate-to-severe depression, especially with the persistent thoughts about not wanting to be alive. The next critical step would be connecting with a therapist who can help create a safety plan and explore treatment options."} showcase that while the sensemaker can provide a potential diagnosis and it fails to explicitly address the patient's susceptibility to their condition and the likelihood of experiencing a health problem due to the underlying condition.\\
In the context of low-scoring perceived severity conversations, the sensemaker identifies the underlying issue and provides a general definition of the condition, along with relating it to the patient's symptoms, but dialogues such as \textit{"Based on your symptoms and recent medication start, I believe you're experiencing akathisia - a restlessness side effect caused by fluphenazine. This can make both relaxation and restful sleep difficult despite not having classic insomnia symptoms."} and \textit{"Based on your symptoms and routine, I believe you may have Delayed Sleep-Wake Phase Disorder, where your internal clock is misaligned with typical day/night cycles. This explains why you can't fall asleep until 2 a.m. and struggle to wake up for school."} shows that the sensemaker does not appropriately convey the seriousness of the potential consequences of the patient's symptoms.\\
Dialogues such as \textit{"Would you like to explore some sleep hygiene strategies that could help you get more restful sleep?"} and \textit{"Using phone reminders and labeled storage containers might help compensate for memory lapses - would you feel comfortable trying those?"} made by sensemaker in the Decision Making stage shows that while the sensemaker provides helpful actions, its strategies might lack details on their efficacy and how these strategies might reduce the risk faced by the patient.

\section{Automated Evaluation}
\label{sec:appendix_automated_conversation_evaluation}

Our three patient-centric communication metrics are inspired by the Health Belief Model and consists of the following dimensions:

\begin{enumerate}[leftmargin=1em]
    \item \textbf{Perceived Susceptibility:} An individual's belief about the likelihood of experiencing a health problem. Table \ref{tab:appendix_perceived_susceptibility_prompt_table} shows the prompt used to evaluate perceived susceptibility.
    \item \textbf{Perceived Severity:} Beliefs about the seriousness of the consequences of a health issue. Table \ref{tab:appendix_perceived_severity_prompt_table} shows the prompt used to evaluate perceived severity.
    \item \textbf{Perceived Benefits:} Beliefs in the efficacy of the advised action to reduce the risk or seriousness of the health impact. Table \ref{tab:appendix_perceived_benefits_prompt_table} shows the prompt used to evaluate perceived severity.
\end{enumerate}

We adapt HBM to evaluate how effectively the sensemaker LLM promotes each of the core dimensions on the patient side during conversations.  
For example, if a patient exhibits low perceived susceptibility or self-efficacy, the sensemaker can adjust the tone and content of messages to address those gaps,  improving engagement and promoting informed decision-making.\\
We evaluate Diagnostic accuracy and conversation flow correctness as well.
\begin{enumerate}[leftmargin=1em]
    \item \textbf{Diagnostic Accuracy:} Since the filtered MedQA~\cite{jin2021disease} case studies provide multi-choice options, we assess two different accuracies: (1) Hard Diagnostic Accuracy, which requires the sensemaker's diagnosis to be an exact match with the ground truth. (2) Soft Diagnostic Accuracy, which allows for some freedom and blanket term matching. Table \ref{tab:appendix_soft_matching_diagnosis_prompt_table} and Table \ref{tab:appendix_hard_matching_diagnosis_prompt_table} show the prompt used to evaluate soft and hard diagnostic accuracy metrics, respectively.
    \item \textbf{Conversation Flow and Correctness:} It is used to measure the ability of the sensemaker to perform end-to-end conversation, while maintaining logical consistency, covering each phase of the conversation, preventing providing incorrect information and preventing contradiction with itself. Table \ref{tab:appendix_flow_correctness_and_logical_consistency_prompt_table} shows the prompt used to evaluate flow correctness.

\end{enumerate}

\begin{table*}[ht]
    \centering
    \small
    \begin{tabular}{p{0.10\linewidth} | p{0.85\linewidth}}
    \toprule
    \textbf{Type} & \textbf{Prompt}\\
    \toprule
         \textbf{System Prompt}
            &Task: You are an expert LLM-as-a-judge model and your task is to evaluate a sensemaker's communication of "Perceived Susceptibility" (definition given as SUSCEPTIBILITY\_DEFINITION) within a patient-sensemaker conversation.\\ 
            &SUSCEPTIBILITY\_DEFINITION: "Perceived Susceptibility" here refers to the sensemaker's ability to clearly and adequately express to the patient their likelihood of experiencing a health problem.\\
            & \\
            &You will be given a conversation between a patient and a sensemaker (given as PATIENT\_SENSEMAKER\_CONVERSATION) and your task is to evaluate the sensemaker's communication of "Perceived Susceptibility" within this conversation based on the following evaluation criteria (given as EVALUATION\_CRITERIA):\\
            & \\
            &EVALUATION\_CRITERIA: Assign a score from 1 (lowest) to 4 (highest) based on the following criteria:\\
            & \\
            & \\
            &1. (Very Poor: Score = 1): The sensemaker failed to communicate any information about the susceptibility of facing a health problem. Additionally, the tone may have been dismissive, insensitive, or unnecessarily alarming. \\
            & \\
            &2. (Poor: Score = 2): The sensemaker failed to communicate any information about the susceptibility of facing a health problem. However, the tone was empathetic and engaging.\\
            & \\
            &3. (Good: Score = 3): The sensemaker communicated basic information about the patient’s susceptibility to facing a health problem. The tone was professional but lacked empathy or collaborative engagement.\\
            & \\
            &4. (Very Good: Score = 4): The sensemaker explained the patient’s susceptibility to facing a health problem clearly, accurately and appropriately, using personalized details effectively. Additionally, the sensemaker was empathetic and engaged collaboratively.\\
            & \\
            &Output Format: You should first provide a short reasoning (2-3 sentences) providing your reasoning for the score. Then, provide the score. Your output should be in the following format:\\
            & \\
            &OUTPUT\_REASONING: <your\_reasoning>\\
            & \\
            &OUTPUT\_SCORE: <score> \\
            & \\
            &Example Output:\\
            & \\
            &OUTPUT\_REASONING: The clinician explicitly and clearly communicated the patient’s susceptibility to depression using tailored information from their personal history. The clinician’s tone was empathetic and engaged the patient in a supportive, collaborative dialogue, ensuring the patient’s clear understanding.\\
            & \\
            &OUTPUT\_SCORE: 4\\
            & \\
            & \\
            &ALWAYS STRICTLY RETURN THE OUTPUT IN THE ABOVE FORMAT.\\
         \midrule
         \textbf{User Prompt}
         &PATIENT\_SENSEMAKER\_CONVERSATION: <Sample conversation>\\
         \bottomrule
    \end{tabular}
    \caption{(Evaluation) Prompt used for automated Perceived Susceptibility evaluation.}
    \label{tab:appendix_perceived_susceptibility_prompt_table}
\end{table*}

\begin{table*}[ht]
    \centering
    \small
    \begin{tabular}{p{0.10\linewidth} | p{0.85\linewidth}}
    \toprule
    \textbf{Type} & \textbf{Prompt}\\
    \toprule
         \textbf{System Prompt}
            &Task: You are an expert LLM-as-a-judge model and your task is to evaluate a sensemaker's communication of "Perceived Severity" (definition given as SEVERITY\_DEFINITION) within a patient-sensemaker conversation. \\
            &SEVERITY\_DEFINITION: "Perceived Severity" refers to the sensemaker's ability to adequately convey the seriousness of the consequences associated with a health issue to the patient.\\
            & \\
            &You will be given a conversation between a patient and a sensemaker (given as PATIENT\_SENSEMAKER\_CONVERSATION) and your task is to evaluate the sensemaker's communication of "Perceived Severity" within this conversation based on the following evaluation criteria (given as EVALUATION\_CRITERIA):\\
            & \\
            &EVALUATION\_CRITERIA: Assign a score from 1 (lowest) to 4 (highest) based on the following criteria:\\
            & \\
            & \\
            &1. (Very Poor: Score = 1): The sensemaker failed to communicate any information about the seriousness of the consequences associated with a health issue to the patient. Additionally, the tone may have been dismissive, insensitive, or unnecessarily alarming.\\
            & \\
            &2. (Poor: Score = 2): The sensemaker failed to communicate any information about the seriousness of the consequences associated with a health issue to the patient. However, the tone was empathetic and engaging.\\
            & \\
            &3. (Good: Score = 3): The sensemaker communicated basic information about the seriousness of the consequences associated with a health issue to the patient. The tone was professional but lacked empathy or collaborative engagement.\\
            & \\
            &4. (Very Good: Score = 4): The sensemaker explained the seriousness of the consequences associated with a health issue clearly, accurately and appropriately, using personalized details effectively. Additionally, the sensemaker was empathetic and engaged collaboratively.\\
            & \\
            &Output Format: You should first provide a short reasoning (2-3 sentences) providing your reasoning for the score. Then, provide the score. Your output should be in the following format:\\
            & \\
            &OUTPUT\_REASONING: <your\_reasoning>\\
            & \\
            &OUTPUT\_SCORE: <score>\\
            & \\
            &Example Output:\\
            & \\
            &OUTPUT\_REASONING: The sensemaker explicitly described the consequences of untreated depression clearly, using detailed personal context relevant to the patient's life situation. The empathetic and collaborative tone ensured patient understanding and minimized unnecessary distress.\\
            & \\
            &OUTPUT\_SCORE: 4\\
            & \\
            & \\
            &ALWAYS STRICTLY RETURN THE OUTPUT IN THE ABOVE FORMAT.\\
         \midrule
         \textbf{User Prompt}
         &PATIENT\_SENSEMAKER\_CONVERSATION: <Sample conversation>\\
         \bottomrule
    \end{tabular}
    \caption{(Evaluation) Prompt used for automated Perceived Severity evaluation.}
    \label{tab:appendix_perceived_severity_prompt_table}
\end{table*}

\begin{table*}[ht]
    \centering
    \small
    \begin{tabular}{p{0.10\linewidth} | p{0.85\linewidth}}
    \toprule
    \textbf{Type} & \textbf{Prompt}\\
    \toprule
         \textbf{System Prompt}
            &Task: You are an expert LLM-as-a-judge model and your task is to evaluate a sensemaker's communication of "Perceived Benefits" (definition given as BENEFITS\_DEFINITION) within a patient-sensemaker conversation. \\
            &BENEFITS\_DEFINITION: "Perceived Benefits" refers to the sensemaker's ability to effectively communicate the efficacy of the advised action in reducing the risk or seriousness of the health impact appropriately.\\
            & \\
            &You will be given a conversation between a patient and a sensemaker (given as PATIENT\_SENSEMAKER\_CONVERSATION) and your task is to evaluate the sensemaker's communication of "Perceived Benefits" within this conversation based on the following evaluation criteria (given as EVALUATION\_CRITERIA):\\
            & \\
            &EVALUATION\_CRITERIA: Assign a score from 1 (lowest) to 4 (highest) based on the following criteria:\\
            & \\
            & \\
            &1. (Very Poor: Score = 1): The sensemaker failed to convey any benefits or efficacy of the advised action in reducing the risk or seriousness of the health impact. Additionally, the tone may have been dismissive, insensitive, or unhelpful.\\
            & \\
            &2. (Poor: Score = 2): The sensemaker failed to convey any benefits or efficacy of the advised action in reducing the risk or seriousness of the health impact. However, the tone was empathetic and engaging.\\
            & \\
            &3. (Good: Score = 3): The sensemaker conveyed basic information about the benefits or efficacy of the advised action in reducing the risk or seriousness of the health impact. The tone was professional but lacked empathy or collaborative engagement.\\
            & \\
            &4. (Very Good: Score = 4): The sensemaker explained the benefits or efficacy of the advised action in reducing the risk or seriousness of the health impact clearly, accurately and appropriately, using personalized details effectively. Additionally, the sensemaker was empathetic and engaged collaboratively.\\
            & \\
            &Output Format: You should first provide a short reasoning (2-3 sentences) providing your reasoning for the score. Then, provide the score. Your output should be in the following format:\\
            & \\
            &OUTPUT\_REASONING: <your\_reasoning>\\
            & \\
            &OUTPUT\_SCORE: <score>\\
            & \\
            &Example Output:\\
            & \\
            &OUTPUT\_REASONING: The sensemaker clearly articulated how therapy sessions could effectively reduce anxiety, using specific examples relevant to the patient’s personal situation. The tone was supportive, empathetic, and collaborative, significantly enhancing the patient's confidence and understanding of the recommended action. \\
            & \\
            &OUTPUT\_SCORE: 4 \\
            & \\
            & \\
            &ALWAYS STRICTLY RETURN THE OUTPUT IN THE ABOVE FORMAT.\\
         \midrule
         \textbf{User Prompt}
         &PATIENT\_SENSEMAKER\_CONVERSATION: <Sample conversation>\\
         \bottomrule
    \end{tabular}
    \caption{(Evaluation) Prompt used for automated Perceived Benefits evaluation.}
    \label{tab:appendix_perceived_benefits_prompt_table}
\end{table*}

\begin{table*}[ht]
    \centering
    \small
    \begin{tabular}{p{0.10\linewidth} | p{0.85\linewidth}}
    \toprule
    \textbf{Type} & \textbf{Prompt}\\
    \toprule
         \textbf{System Prompt}
            &Task: You are an expert LLM-as-a-judge model and your task is to evaluate a sensemaker's "Diagnostic Accuracy" (definition given as DIAGNOSTIC\_ACCURACY\_DEFINITION).\\
            &DIAGNOSTIC\_ACCURACY\_DEFINITION: "Diagnostic Accuracy" refers to the accuracy of the sensemaker's diagnosis when compared to the ground truth diagnosis.\\
            &You will be given a specific part from a patient-sensemaker conversation, the sensemaker's diagnosis statement (given as PATIENT\_DIAGNOSIS) and the ground truth diagnosis of the patient (given as GT\_DIAGNOSIS). Your task is to evaluate the sensemaker's "Diagnostic Accuracy" from this conversation snippet and assign a binary label based on the following evaluation criteria (given as EVALUATION\_CRITERIA):\\
            & \\
            &EVALUATION\_CRITERIA:\\
            &1. Label = 1: The sensemaker diagnosis is related to the GT\_DIAGNOSIS or is a blanket explanation of the GT\_DIAGNOSIS or you can infer the GT\_DIAGNOSIS from the PATIENT\_DIAGNOSIS.\\
            &2. Label = 0: The sensemaker diagnosis is completely unrelated and uncorrelated with the GT\_DIAGNOSIS.\\
            & \\
            &Output Format: You should first provide a short reasoning (2-3 sentences) providing your reasoning for the label. Then, provide the label. Your output should be in the following format:\\
            &OUTPUT\_REASONING: <your\_reasoning>\\
            &OUTPUT\_SCORE: <0/1> \\
            & \\
            &Example Output:\\
            &OUTPUT\_REASONING: The sensemaker diagnosed the person with anxiety while the ground truth diagnosis is PTSD, since they are both are closely related conditions and PTSD can be inferred from the patient diagnosis snippet of the conversation, therefore the label is 1.\\
            &OUTPUT\_SCORE: 1\\
            & \\
            & \\
            &ALWAYS STRICTLY RETURN THE OUTPUT IN THE ABOVE FORMAT.\\
         \midrule
         \textbf{User Prompt}
         &PATIENT\_DIAGNOSIS: <Providing Information stage Message>\\
         &GT\_DIAGNOSIS: <Ground Truth Label>\\
         \bottomrule
    \end{tabular}
    \caption{(Evaluation) Prompt used for hard matching diagnostic accuracy}
    \label{tab:appendix_soft_matching_diagnosis_prompt_table}
\end{table*}

\begin{table*}[ht]
    \centering
    \small
    \begin{tabular}{p{0.10\linewidth} | p{0.85\linewidth}}
    \toprule
    \textbf{Type} & \textbf{Prompt}\\
    \toprule
         \textbf{System Prompt}
            &Task: Task: You are an expert LLM-as-a-judge model and your task is to evaluate a sensemaker's "Diagnostic Accuracy" (definition given as DIAGNOSTIC\_ACCURACY\_DEFINITION). \\
            &DIAGNOSTIC\_ACCURACY\_DEFINITION: "Diagnostic Accuracy" refers to the accuracy of the sensemaker's diagnosis when compared to the ground truth diagnosis.\\
            &You will be given a specific part from a patient-sensemaker conversation, the sensemaker's diagnosis statement (given as PATIENT\_DIAGNOSIS) and the ground truth diagnosis of the patient (given as GT\_DIAGNOSIS). Your task is to evaluate the sensemaker's "Diagnostic Accuracy" from this conversation snippet and assign a binary label based on the following evaluation criteria (given as EVALUATION\_CRITERIA):\\
            & \\
            &EVALUATION\_CRITERIA:\\
            &1. Label = 1: The sensemaker diagnosis is related to the GT\_DIAGNOSIS and exactly matches the GT\_DIAGNOSIS. \\
            &2. Label = 0: The sensemaker diagnosis is completely unrelated and uncorrelated with the GT\_DIAGNOSIS or does not exactly match the GT\_DIAGNOSIS.\\
            & \\
            &Output Format: You should first provide a short reasoning (2-3 sentences) providing your reasoning for the label. Then, provide the label. Your output should be in the following format:\\
            &OUTPUT\_REASONING: <your\_reasoning>\\
            &OUTPUT\_SCORE: <0/1>\\
            & \\
            &Example Output:\\
            &OUTPUT\_REASONING: The sensemaker diagnosed the person with anxiety while the ground truth diagnosis is PTSD, since they are not exactly the same, therefore the label is 0.\\
            &OUTPUT\_SCORE: 0\\
            & \\
            & \\
            &ALWAYS STRICTLY RETURN THE OUTPUT IN THE ABOVE FORMAT.\\
         \midrule
         \textbf{User Prompt}
         &PATIENT\_DIAGNOSIS: <Providing Information stage Message>\\
         &GT\_DIAGNOSIS: <Ground Truth Label>\\
         \bottomrule
    \end{tabular}
    \caption{(Evaluation) Prompt used for hard matching diagnostic accuracy.}
    \label{tab:appendix_hard_matching_diagnosis_prompt_table}
\end{table*}

\begin{table*}[ht]
    \centering
    \small
    \begin{tabular}{p{0.10\linewidth} | p{0.85\linewidth}}
    \toprule
    \textbf{Type} & \textbf{Prompt}\\
    \toprule
         \textbf{System Prompt}
            &Task: You are an expert LLM-as-a-judge model and your task is to evaluate a sensemaker's *Conversation Flow and Correctness* (definition given as FLOW\_CORRECTNESS\_DEFINITION) within a patient-sensemaker conversation. \\
            &FLOW\_CORRECTNESS\_DEFINITION: "Conversation Flow and Correctness" refers to the sensemaker’s ability to (i) progress through an end-to-end sensemaking dialogue in a logically coherent order according to STAGE\_DEFINITION order and (ii) provide consistent information without contradictions or inconsistencies.\\
            & \\
            &You will be given a conversation between a patient and a sensemaker (given as PATIENT\_SENSEMAKER\_CONVERSATION) and your task is to evaluate the sensemaker's performance on "Conversation Flow and Correctness" based on the following evaluation criteria (given as EVALUATION\_CRITERIA):\\
            & \\
            &EVALUATION\_CRITERIA: Assign a score from 1 (lowest) to 4 (highest) using the rubric below:\\
            & \\
            &1. (Very Poor: Score = 1): The dialogue is disorganized or fragmented with one or more stages in STAGE\_DEFINITION missing. Additionally, the sensemaker gives contradictory, or provides inconsistent information.\\
            & \\
            &2. (Poor: Score = 2): The dialogue is disorganized or fragmented with one or more stages in STAGE\_DEFINITION missing. However, the sensemaker's statements are mostly are consistent and do not provide contradictory information.\\
            & \\
            &3. (Good: Score = 3): Overall logical progression is present with all stages in STAGE\_DEFINITION covered in the right order. Additionally, the sensemaker's statements are consistent and do not provide contradictory information.\\
            & \\
            &4. (Very Good: Score = 4): Overall logical progression is present with all stages in STAGE\_DEFINITION covered in the right order and the sensemaker has covered the each stage sufficiently paying special attention on gathering information from the patient and providing appropriate information and empathetic support to the patient. Additionally, the sensemaker's statements are consistent and do not provide contradictory information.\\
            & \\
            &\#\# STAGE\_DEFINITION\\
            &A typical sensemaking conversation between a sensemaker and a patient consists of the following order of stages:\\
            & \\
            &- Fostering the Relationship: Establish a trusting, open, and collaborative rapport with the patient by demonstrating empathy, honesty, and respect.\\
            &- Gathering Information: Develop a thorough understanding of the patient’s condition by exploring both biological and psychosocial factors, including their needs, concerns, and medical history.\\
            &- Providing Information: Offer a potential diagnosis that directly addresses the patient's concerns, ensuring clarity and sensitivity in communication.\\
            &- Decision Making: Respond to the patient’s medical questions regarding the diagnosis, and recommend only lifestyle or non-clinical interventions to help manage their condition.\\
            &- Responding to Emotions: Acknowledge and address the emotional aspects of the patient’s experience by expressing empathy, offering reassurance, and providing appropriate psychological support.\\
            &- Exit: Conclude the conversation in a respectful and supportive manner, ensuring the patient feels heard and understood.\\
            & \\
            &\#\# OUTPUT FORMAT\\
            &Provide a brief reasoning (2-3 sentences) explaining the score, followed by the score itself, using the template below:\\
            &OUTPUT\_REASONING: <your\_reasoning>\\
            & \\
            &OUTPUT\_SCORE: <score>\\
            & \\
            &Example Output:\\
            & \\
            &OUTPUT\_REASONING: The sensemaker gathered a complete history, summarized findings, explained differential diagnoses, and covered all stages in STAGE\_DEFINITION in the right order with appropriate information and empathetic support to the patient. No contradictions or inconsistencies were detected.\\
            & \\
            &OUTPUT\_SCORE: 4\\
            & \\
            &ALWAYS STRICTLY RETURN THE OUTPUT IN THE ABOVE FORMAT.\\
         \midrule
         \textbf{User Prompt}
         &PATIENT\_SENSEMAKER\_CONVERSATION: <Sample conversation>\\
         \bottomrule
    \end{tabular}
    \caption{(Evaluation) Prompt used for Flow Correctness and Logical Consistency evaluation.}
    \label{tab:appendix_flow_correctness_and_logical_consistency_prompt_table}
\end{table*}

\section{Human Annotators and Annotation Guidelines}
\label{appendix_sec:human_annotation_details}

Human annotations was conducted by four graduate students fluent in English language and with a background in Computer Science and digital mental health research. For the annotation task,annotators were provided with the entire patient-sensemaker conversation, score for each metric (\emph{Perceived Susceptibility, Perceived Severity, Perceived Benefits and Conversation Flow and Correctness, Diagnostic accuracy}) from the automated evaluation, reasoning for the score generated by LLM judge and the rubric with the definitions for each metric and scoring criteria.

For each metric and conversation, two annotators assigned a binary label indicating agreement or disagreement with the LLM judge’s score and reasoning for the metric. While a positive label denoted agreement with the LLM score and reasoning, a negative label indicated non-alignment with LLM judge score or reasoning. Overall, we observed high correlation for each metric.  For \emph{Diagnostic Accuracy (Hard)}, the average correlation with LLM judge was computed as 93.5\%, whereas as for \emph{Diagnostic Accuracy (Soft)}, average correlation was 90\%. For the patient-centric communication metrics, agreement scores were 79\% for \emph{Perceived Susceptibility}, 87.5\% for \emph{Perceived Severity}, 86.5\% for \emph{Perceived Benefits}. Finally, \emph{Conversation Flow and Correctness} average correlation with LLM judge was computed as 87\%.

\section{Model Details, Hyperparameters, and Compute}
\label{appendix_sec:compute_details}
We use API-based model inference for GPT-4o, OpenAI o1, and DeepSeek R1. We used Azure OpenAI Service for accessing GPT-4o,  OpenAI o1, and DeepSeek R1 models. 
The hyperparameters for the generations are presented below. Table \ref{tab:appendx_model_details} presents the details regarding the model sizes and versions.\\
\begin{table*}[!h]
\centering

\begin{tabular}{|l|l|l|}
\hline
\textbf{LLM} & \textbf{Version} & \textbf{Parameter Size} \\ \hline
GPT-4o & \cite{gpt4o} [Version: 2024-11-20] & (undisclosed) \\ \hline
OpenAI o1 & \cite{openaio1} [Version: 2024-12-17] & (undisclosed) \\ \hline
DeepSeek R1 & \cite{guo2025deepseek} [Version: 1] & 671 billion\\ \hline
\end{tabular}
\caption{Models and their corresponding version dates}
\label{tab:appendx_model_details}
\end{table*}

\noindent\textbf{Hyperparameters for GPT and DeepSeek Models}: temperature $t=0$, max\_tokens $=1000$ (for sensemaker fact extraction, matching and message redundancy checker which use GPT-4o) and max\_tokens $=15000$ (for
sensemaker next stage determination and message generation which use reasoning models). All patient module components use GPT-4o with a temperature $t=0$ and max\_tokens $=400$.

\section{Information About Use of AI Assistants}
\label{appendix_sec:ai_use_info}

We used AI assistants for text rephrasing. The usage was limited to correcting grammatical mistakes and choice of words.

\end{document}